\documentclass[9pt,technote]{IEEEtran}

\IEEEoverridecommandlockouts                              

\usepackage{placeins}
\usepackage{algorithm}
\usepackage{algorithmic}
\usepackage{graphicx}
\usepackage{wrapfig}
\usepackage[bookmarks=false]{hyperref}
\usepackage{flushend}
\usepackage{multirow}
\usepackage{caption}
\usepackage{subcaption}
\usepackage{amsmath}
\usepackage{amssymb}

\usepackage{tabularx, booktabs}

\usepackage{mathtools, cuted}
\usepackage{lipsum, color}

\usepackage{mwe,tikz}
\usepackage[percent]{overpic}

\usepackage{tikz}
\usetikzlibrary{shapes,arrows, calc}
\usepgflibrary{arrows}

\usepackage{algorithm}
\usepackage{algorithmic}
\usepackage{graphicx}
\usepackage{wrapfig}
\usepackage[bookmarks=false]{hyperref}
\usepackage{flushend}
\usepackage{multirow}
\usepackage{graphicx}
\usepackage{caption}
\usepackage{subcaption}
\usepackage{amsmath}
\usepackage{amssymb}

\usepackage{mathtools, cuted}
\usepackage{lipsum, color}

\usepackage{tabularx, booktabs}
\usepackage{pbox}
\usepackage{mwe,tikz}
\usepackage[percent]{overpic}

\usepackage{tikz}
\usetikzlibrary{shapes,arrows,calc,positioning}
\usepgflibrary{arrows}

\definecolor{arrowblue}{RGB}{98,145,224}

\DeclareMathOperator*{\argmin}{arg\,min}
\DeclareMathOperator*{\argmax}{arg\,max}

\title{\LARGE \bf
Adaptive Cost Function for Pointcloud Registration
}

\author{Johan Ekekrantz$^{*}$, John Folkesson and Patric Jensfelt
\thanks{$^{*}$ Corresponding author. }
\thanks{The authors are with the Centre for Autonomous System at KTH Royal Institute of Technology, SE-100 44 Stockholm, Sweden
        {\tt\small \{ekz,johnf,patric\}@csc.kth.se}}%
\thanks{The work presented in this papers has been funded by the European Union Seventh Framework Programme (FP7/2007- 2013) under grant agreement No 600623 (”STRANDS”), the Swedish Foundation for Strategic Research (SSF) through its Centre for Autonomous Systems and the Swedish Research Council (VR) under grant C0475401.}%
}

\begin{document}
\maketitle
\thispagestyle{empty}
\pagestyle{empty}

\begin{abstract}

In this paper we introduce an adaptive cost-function for pointcloud registration. The algorithm automatically estimates the sensor noise, which is important for generalization across different sensors and environments. Through experiments on real and synthetic data, we show significant improvements in accuracy and robustness over state-of-the-art solutions.

\end{abstract}


\section{INTRODUCTION}

\begin{figure*}
    \begin{subfigure}{0.49\textwidth}
        \includegraphics[width=\textwidth]{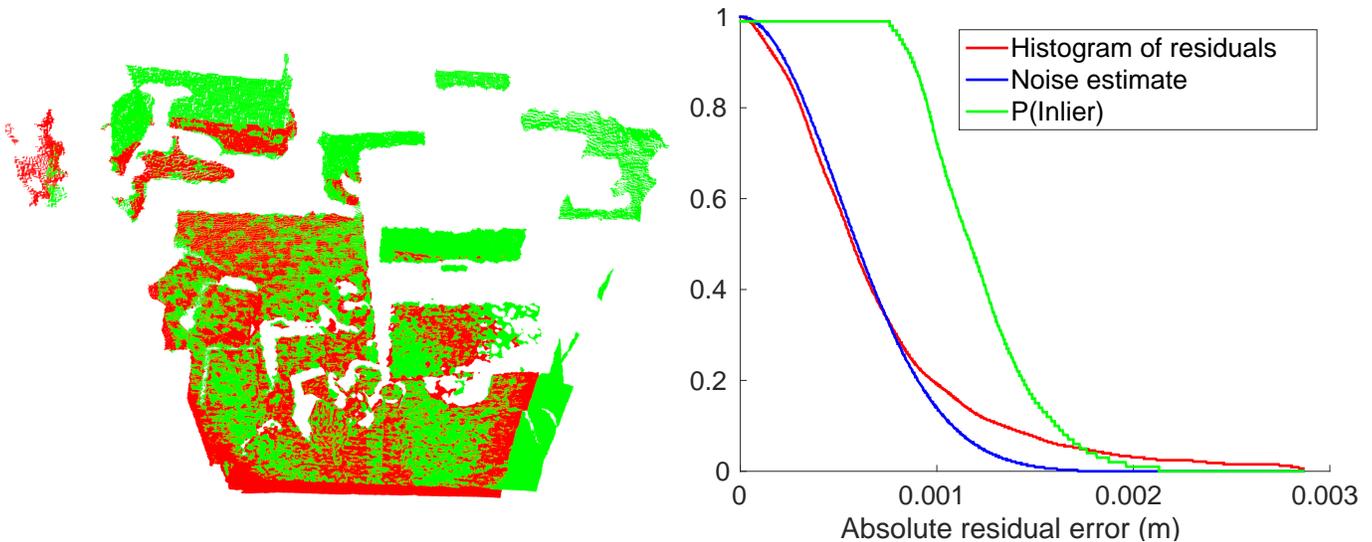}
        \label{fig:registration}
    \end{subfigure}
    \hfil
    \begin{subfigure}{0.49\textwidth}
        \includegraphics[trim={1 3 7 5},clip,width=\textwidth]{imgs/regplot}
        \label{fig:inliersTransformedOutliers}
    \end{subfigure}

   \caption{Left: Two point-clouds registered using our method. Right: Resulting histogram of residuals error (inliers and outliers), noise estimate (inliers) and probability of inliers as function of residual error. The non-overlapping outlier regions of the point-clouds appear at the tail end of the histogram of residuals.}
    \label{fig:registrationExample}
\end{figure*}


\IEEEPARstart{T}{he} task of aligning sensor data is called registration and is an
important field in robotics, with applications such as simultaneous
localization and mapping, object tracking and 3D reconstruction. For
two point-clouds, registration is equivalent to finding a relative
sensor pose transformation.  The modelling of sensor noise can make
registration more robust and accurate. Our approach adaptively learns
the sensor noise model using the measured point-cloud data while performing the
registration.

Registration problems are usually solved by finding corresponding points in
different point-clouds and from these correspondences estimate a
transformation between the point-clouds. Unfortunately, there is no
universal solution, which results in only
correct correspondences, for the point matching problem.  Many popular solutions require
sensor-specific tuning of parameters to reduce the influence of
incorrect correspondences, such parameters make using the registration
algorithm very complex for non-expert users. It is therefore desirable
to find solutions which easily port between different sensors and
platforms. In practice, many factors such as illumination strength or
environment temperature may affect the sensor noise model. Such
environment variables are hard to account for beforehand, making
precise a priori noise models impractical in many scenarios. A
solution to these problems is to estimate the sensor noise
model on the fly, directly from measurement data.

Often when point-cloud registration is required, a coarse initial guess
of the alignment is already available. Such guesses can come from
robot odometry or from the fact that the sensor only can move so far
in a certain amount of time. This class of problems is called
constrained registration. If no initial guess is available, it is
called unconstrained registration. In the case of constrained
registration, solutions usually perform iterative refinement of the
correspondences and the estimated transform between the
point-clouds. For constrained registration, the range of convergence on
the accuracy of the initial guess is an important factor.

An additional issue can be that the sensors discretely sample the
environment, further complicating the registration.  This is not
addressed by our method.  In this paper we limit the problem to
constrained registration in which the separation of inliers from
outliers is the primary source of error.  In particular, we assume
that the two point clouds that are to be matched contain some common
points that are sampled from nearly the same 3D point in the
environment and some that are outliers, ie. points in one point cloud
that have no corresponding points in the other point cloud. This is the same assumption made by most registration algorithms.

The main contribution of this paper is a method which we call
Statistical Inlier Estimation (SIE). SIE can be used to analyze a set
of pairwise matched measurements and from that compute a probability
for every match to be correct. The basic idea for SIE is to model the
distribution of measurements using histograms of residual  errors, and from
these histograms estimate the distribution of inliers.  The residuals
are computed using an estimated transformation between the
point-clouds. The estimation of the inlier distribution and the
transformation are done alternatively and iteratively.  Having an estimate
of the inlier probability allows us to define a cost function that does
soft matching rather than binary. That is, rather than completely removing points classified as outliers
from the cost, all closest matched point pairs are included in the
cost but with a weight that depends on the inlier probability.

The registration can be performed without any requirement on hand
tuned parameters for specific sensors or applications. From a practical perspective, SIE is relatively
simple to implement.  The same idea
has previously~\cite{ppr} been used for fitting surface primitives
such as planes and spheres to point-cloud data. In this paper we show
that it can be used as a general purpose inlier estimation
framework. We extend the framework to allow for arbitrary noise
distributions and enable 3D point-to-point inlier estimation as
opposed to only point-to-surface in~\cite{ppr}. We also adapt the method
by which the distribution of inliers is estimated. We show that SIE
can be used to extend the standard \emph{ICP}~\cite{Besl92} algorithm.
By adding an adaptively decreasing regularization to the variance of
our estimated noise model we improve convergence. The regularization
approximately correspond to the current uncertainty of the estimated 
parameters in each iteration of the ICP algorithm. We compare ICP
registration using SIE to other methods on both simulated and real
world data.

\section{RELATED WORKS}

By far the most popular solution to the constrained point-cloud
registration problem is the Iterative Closest Point~\emph{ICP}
algorithm~\cite{Besl92}. ICP works
directly on the point-clouds by alternating between performing point
matching using point-to-point nearest neighbour matching and
recomputing the relative transform between the two frames such that it
minimizes the L2-norm of the matches. If the sensor measurement noise
is Gaussian, the L2-norm correspond to the maximum likelihood estimate
of the transform.

Using the assumption that points are sampled from continuous surfaces, several refinements have been proposed to remove the discretization noise generated by the point sampling from a surface. In \cite{Yang92} point-to-plane \emph{ICP} is introduced, meaning that the point distances in one cloud are minimized as the projection distance along the normals of the matched points. In \cite{gicp} the GICP algorithm is introduced. GICP can be seen as plane-to-plane ICP. In~\cite{NICP}, the NICP algorithm was introduced. During data association, NICP accounts for the relative surface orientations of the points and the local curvature, improving the matching and therefore the accuracy and robustness of the estimation. NICP also change the optimization criteria to minimize the total Mahalanobis distance of the found correspondence
point pairs and their normals. In~\cite{colorNICP} the NICP algorithm was extended to use color information to improve the point matching. 

In~\cite{trimmedicp} the trimmed \emph{ICP} algorithm is proposed. To cope with outliers, correspondences with residuals outside of some chosen quantile of the residuals for all correspondences are rejected. This can be a good option if the number of outliers is approximately known a priori.

In~\cite{lmICP}, the authors propose to use Levenberg-Marquardt algorithm to perform the estimation of the transformation given point correspondences. The Levenberg-Marquardt algorithm is a general-purpose non-linear optimization method, allowing for arbitrary cost-functions to be used.

In~\cite{multiviewLMICP} a multi-view Levenberg-Marquardt ICP algorithm is introduced. The algorithm estimates the scale of the sensor noise using the median residual. Using the median residual to estimate the approximate size of the sensor noise removes the need for a hand-tuned rejection criteria, but assumes that atleast half of the residuals come from inlier correspondences and has a built-in bias which increases as the number of outliers grow. The median residual value has some similarity to our use of an adaptively decreasing regularization in that the estimated noise often is larger at the start of the ICP optimization than at the end.

In~\cite{trustICP} trust region optimization was used to perform non-linear optimization of robust cost-functions for ICP. Similarly to~\cite{multiviewLMICP} the approximate size of the sensor noise is estimated using the median residual.

In GO-ICP~\cite{goicp16}, a branch-and-bound and \emph{ICP} hybrid algorithm, is introduced. The algorithm is proven to find the optimal solution up to a chosen level of accuracy for the point-to-point L2 error metric and a trimmed \emph{ICP}~\cite{trimmedicp}. Unfortunately the increased range of convergence comes at a price of significantly increased computational complexity.

In~\cite{sparseicp_sgp13}, the authors propose the usage of L$p$-norms where $p < 2$ in order to better cope with false correspondences. This work is similar to ours in that it does not require complicated tuning for use on a new sensor.

When using a kd-tree (or a set of trees), the nearest neighbour matching of \emph{ICP} has the computational complexity of $\mathcal{O}(Nlog(K))$, where $N$ is the number of points being matched and $K$ is the number of points matched against. Another popular alternative for range scans is to use back projection in order to find correspondences. While back projection does not guarantee to find the true nearest neighbour match, the matching step can be done in $\mathcal{O}(N)$, providing a significant reduction in computational complexity. The registration component of popular 3D-SLAM systems such~\cite{kerl13iros} and ~\cite{newcombe2011kinectfusion} that require 30 frames per second performance use back projection as a matching algorithm. ~\cite{kerl13iros} uses an automatically estimated T-distribution for estimation.



Image registration has traditionally been done using keypoints, a subset of the pixels in the image. Common keypoint extractors such as FAST~\cite{rosten2006machine} or Harris corners~\cite{Harris88acombined} can find a stable subset of points in images quickly. A large body of work has been put into finding distinctive descriptors for visual information, such as SIFT~\cite{lowe2004distinctive}, SURF\cite{bay2006surf}, BRIEF~\cite{calonder2010brief}, ORB~\cite{rublee2011orb} and GRIEF~\cite{grief2}. Keypoint extractors and descriptors have also been created for point-cloud data, such as 3D-sift~\cite{sift3d}, NARF~\cite{narf} and SHOT~\cite{shot}.

In addition to descriptor similarity, geometrical constraints can be used to filter out false correspondences. Probably the most popular technique used for this is the RANSAC~\cite{RANSAC} algorithm. In~\cite{ekekrantz2013ecmr} and~\cite{Ekekrantz13iros} the AICK algorithm was introduced. AICK is an \emph{ICP}-like algorithm which quickly registers two sets of 3D points with associated descriptors. AICK use a weighted distance metric, taking into account both feature similarity of keypoints and geometrical fit. AICK incrementally change the importance of the feature similarity between the keypoints and the geometrical fit as registration is performed. This relaxes the requirement on geometrical fit at the beginning of the registration, gradually increasing the importance of the geometrical fit for both the matching and outlier rejection as the registration improves from iteration.

In~\cite{fastGlobalRegistration} a keypoint-based registration technique which uses a scaled Geman-McClure estimator, where the $\lambda$ parameter is sequentially decreased as the registration is improved. This gradually increases the importance of the geometrical fit for determining the influence of a match based on the geometrical fit as the registration improves.

~\cite{angle_histograms} presents a pointcloud registration method which forms histograms over surface orientations. The surface orientation histograms are then directly aligned to register the pointclouds. This technique efficiently takes advantage of the high amount of co-planar surface patches in indoor environments, leading to improved robustness to poor initialization estimates.

The normal distribution transform (NDT) was developed to take sensor noise into account, first in 2D~\cite{Biber03} and later extended to 3D in~\cite{Magnusson07,Stoyanov_etal:ICRA:2012}. These methods represent one of the point-clouds using normal distributions to which the points in the other point-clouds are aligned. In~\cite{d2d-NDT} the technique is modified to perform registration from a set of normal distributions to another set of normal distributions. In~\cite{huhle2008fly} 3D-NDT was augmented with matched keypoints. The weight between the matched keypoints and the NDT results was controlled by a weight that changed over iterations based on the fit of the NDT. In~\cite{huhle2008registration} 3D-NDT was extended to use color information.

In~\cite{ppr} an algorithm for fitting geometric primitives to point-cloud data was introduced. At the core of the algorithm is an inlier estimation system which is specifically created for Gaussian distributed point-to-surface distances. ~\cite{ppr} and SIE share the idea of fitting a noise distribution to a distance-histogram. Compared to~\cite{ppr}, SIE allows for any distribution to be used. The application domain also differs, as well as many of the practical details as to how the distributions are fitted. SIE also allows for multi-dimensional residuals, which~\cite{ppr} does not.

\section{Point-cloud Registration}
\label{section:Registration}
Point-cloud registration is done by maximum
likelihood estimation over the relative positions of the
point-clouds. This finds the transformation that
minimizes some cost function over a set of residual errors.
In this paper we will assume that the sensor measurement noise
distribution is Gaussian. Using other distributions require minimal changes, see appendix for details and experiments using non-Gaussian noise.

Given two point-clouds $A = \{a_0,a_1,\dots,a_n\}$ and $B =
\{b_0,b_1,\dots,b_m\}$ and assuming some correspondence map between
them, the probability of residual $r_{i} = T*a_i-b_j$ is modelled
\begin{equation}
\label{eq:point2surfaceprob}
P(a_i|b_j,T) = \mathcal{N}(||r_{i}||,0,\sigma_i)
\end{equation}
The registration problem can therefore be formulated as the computation of $\argmax_{b,T} P(A|b,T)$, where we indicate by $b$ the correspondences from $B$ that are matched to $A$. We can reformulate the optimization problem as
\begin{equation}
\label{eq:regprob}
\begin{split}
& \argmin_{b,T} -log \Big( P(A|b,T) \Big) = \argmin_{b,T} \sum_{i=0}^{n}(\frac{||r_i||}{\sigma_i})^2
\end{split}
\end{equation}

The Iterative Closest Point, \emph{ICP}, algorithm assumes that $T$ is approximately known beforehand.   The maximum likelihood estimate of $b_i$ is the closest point to $T*a_i$. We will refer to the first step of ICP, where nearest neighbour matching between the point-clouds is performed, as the matching step. The second step of the \emph{ICP} algorithm where eq.(\ref{eq:regprob}) is minimized by refining the estimate of $T$ given the previously found point-to-point correspondences, will be referred to as the refinement step. Given the correspondences $b$, minimization of eq.(\ref{eq:regprob}) for the $T$ variable can be performed using a standard weighted least squares minimization.

\begin{equation}
\label{eq:regprob2}
\begin{split}
& T_{k+1}= \argmin_{T} \sum_{i=0}^{n}w(r_i^{k})||r_i||^2
\end{split}
\end{equation}

Using the Iteratively Re-weighted Least Squares (IRLS) algorithm, robust cost functions can be used to reduce the influence of outliers by iteratively solving a set weighted least squares minimizations.
 
IRLS is performed using eq.(\ref{eq:regprob2}) where k indicates the iteration of the IRLS and the superscripts
indicate the $T$ that the residuals were computed with. We will
compare various standard choices for $w(i)$.  Some of these are
designed to give lower or even zero weight to residuals that are less
likely to be inliers.  We will show in the next section how we can
approximate $P(I|r_i)$, the probability that a residual is an inlier
given its value.  In our method, the cost function is given a weight proportional to $P(I|r_i)$:

\begin{equation}
\label{eq:regprob3}
\begin{split}
w(r_i) = \sigma_i^{-2} * P(I|r_i)
\end{split}
\end{equation}

\section{Statistical Inlier Estimation (SIE)}
\label{section:inlierestimation}

The ability to detect or mitigate the effect of outliers is critical
for the accuracy of point-cloud registration. Even a very small
fraction of unaccounted for outliers generally results in extremely
poor estimations.

SIE builds a histogram over the residual errors, $r_i$, and uses that histogram to
estimate the parameters of the measurement noise model. The residuals
are the differences between the matched point pairs using the current
best estimate of the transformation between point clouds.
For estimating the measurement noise model, we exploit the insight
that, at or near the peak in the histogram, the contribution to the histogram of
outliers is small as compared to that of the inliers.

Fitting a parametric noise model to the histogram shape near the peak
therefore models the inlier distribution, with negligible effect from the
outliers. Since the histogram is the sum of the outliers and the
inliers, the ratio of the inlier distribution to the histogram tells
us about the outlier versus inlier relative probability. Figure~\ref{fig:registrationExample} shows the estimated distributions and probabilities of SIE for a registered pointcloud. For the
registration, we use this to weight all correspondence pairs in the
cost function. This cost function is then minimized to give a new
transformation (the weights are held constant during this
minimization).  We can then iterate this process with new residuals.
\subsection{Adaptive model fitting}
\label{subsection:trainingstep}

The adaptive model fitting step takes as input a set of residuals, in the form of a matrix $R$ with $n$ rows and $m$ columns, $n$ being the number of correspondences and $m$ the number of dimensions for the residuals caused by the correspondences. 

For each column $j$ of $R$, we compute a histogram $H_j$.  Each histogram is then the sum of the distributions of the inliers and the outliers. The distribution of outliers is in general unknown and often of non-parametric form and is as such hard to compute directly from the data or define a priori. The inlier measurement distribution on the other hand can be  approximated by some a-priori known parametric function, making it well suited for estimation. In practice, normal (Gaussian) distributions are very popular approximations of sensor noise for a wide variety of sensor types.

In section~\ref{section:Registration} we made the assumption that the
sensor noise, and therefore the inliers, were distributed according to
$\mathcal{G}_j =
\alpha_j*e^{-0.5|\tfrac{x-\mu_j}{\sigma_j}|^{2}}$.  We make the
assumption that all residuals within one column of $R$ are independent
and identically distributed.  In many applications the sensor noise
model parameters (here $\{ \alpha_j,\mu_j,\sigma_j \}$) are
unknown and impossible to define a priori, due to among other things
environmental conditions. Cameras for example are susceptible to
environmental conditions such as illumination, unknown camera exposure
time or temperature changes. We believe that if the parameters are
unknown or at least inaccurately known, the parameters should be
estimated from data.

The $\mu_j$ value is defined by a clear peak in $H_j$, see
fig.(\ref{fig:registrationExample}) for example. Assuming that the
vast majority of measurements at the peak correspond to inliers, we
can draw the conclusion that $\alpha_j \approx H_j(\mu_j)$. Once
$\alpha_j$ and $\mu_j$ are approximately known, we estimate $\sigma_j$
by fitting the noise estimate $\mathcal{G}_j$ to the
histogram $H_j$. $\mathcal{G}_j$ is fitted by solving
eq.(\ref{eq:fitParams}), which corresponds to maximizing the data
likelihood under the soft constraint that $G_j \leq H_j$. As such, we
seek a sensor noise model estimate $\mathcal{G}_j$ that explain as
many measurements as possible without violating probability
constraints.

\begin{equation}
\label{eq:fitParams}
\begin{split}
\argmin_{\sigma_j} \sum_{i=0}^{n} F\Big( H_j(i)-\mathcal{G}(i,\alpha_j,\mu_j,\sigma_j) \Big) \\
 F(x) = 
  \begin{cases} 
   -k * x & \text{if } x \leq 0 \\
   x  & \text{if } x > 0
  \end{cases}
\end{split}
\end{equation}
where $k$ is used as a penalty term that push the minimization to choose solutions where $H_j(i)\geq \mathcal{G}(i,\alpha_j,\mu_j,\sigma_j)$. 

We found that eq.(\ref{eq:fitParams}) can be effectively minimized using bisection, but any other suitable optimization technique would also be fine. 

Computing the inlier distribution parameters in the histogram space has the advantage that, other than constructing the histogram, the computational cost is not related to the number of samples but rather the number of bins in the histogram. This keeps the computational complexity of the minimization low even if there are vast quantities of fitting data and a complex inlier distribution.

\subsection{Prediction step}
\label{subsection:predictionstep}
Once the inlier noise distributions and the histograms are known, one
can compute the inlier probability for a single dimension $j$ of a
measurement $i$.  We start with the component of the residual
distribution that is due to the inliers,
\begin{equation}
P(I|R_{i,j})P(R_{i,j})=P(R_{i,j}|I)P(I)\approx\mathcal{G}_j(R_{i,j})
\end{equation}

where P(I) is the prior probability of inliers, and $P(R_{i,j})$ is
given by the histogram. One way to estimate P(I) is by summing over the data:

\begin{equation}
\label{eq:pi}
P(I)\approx \sum_{i,j}\mathcal{G}(R_{i,j})
\end{equation}

We will throughout implicitly assume that we have normalized $H_j$ and $\mathcal{G}_j(R_{i,j})$ so that
$\sum_{i,j}H_j(R_{i,j})=1$, even if in many formulas this normalization
factor cancels and need not be computed. For one dimension the inlier 
probability is the ratio of expected inliers
to the ratio of total residuals for a specific bin in the histogram.

\begin{equation}
\label{eq:InlierRatio}
P(I|R_{i,j}) \approx \frac{ \mathcal{G}(R_{i,j},\alpha_j,\mu_j,\sigma_j)}{ H_j(R_{i,j}) }
\end{equation}

We know that a single correspondence, a row $i$ of $R$, is either an
inlier or outlier.  We have computed the probability of it being an
inlier given one column of that row.  We now have to compute the
probability given the evidence from all the columns of the row.  The
residuals across a row are not independent unless we know whether or
not the row is an inlier.  We need to find the joint probability across
all dimensions $j$.

\begin{equation}
P(R_{i})= P(I)\prod_j P(R_{i,j}|I) + P(O)\prod_j P(R_{i,j}|O)
\end{equation}

where $O$ indicates outlier, $P(O)=1-P(I)$. 

\begin{equation}
P(R_{i,j}|O)\approx \frac{P(R_{i,j})-P(R_{i,j}|I)P(I)}{P(O)}
\end{equation}

In terms of our estimated inlier distribution and the histogram:
\begin{equation}
P(R_{i,j}|O)\approx \frac{H_j(R_{i,j})-\mathcal{G}(R_{i,j})}{P(O)}
\end{equation}
And finally we can infer the inlier probability

\begin{equation}
P(I|R_{i})\approx \frac{\prod_j\mathcal{G}(R_{i,j})}{\prod_j\mathcal{G}(R_{i,j})+(P(I))^{m-1}P(O)\prod_jP(R_{i,j}|O)}
\end{equation}

This can be rewritten as

\begin{equation}
P(I|R_{i})\approx \frac{\prod_jP(I|R_{i,j}) }{\prod_jP(I|R_{i,j}) +\gamma\prod_j(1-P(I|R_{i,j})) }
\end{equation}

where $\gamma=(P(I)/P(O))^{m-1}$.  This formula is the inlier
probability given the $m$ dimensional residual. To avoid degenerate cases leading to division by zero, we truncate the value of $P(I|R_{i,j})$ to some value less then 1 (in this paper we pick 0.99). This means that, regardless of residual value, no correspondence is completely certain to be correct.

\subsection{Regularizer for use in parameter estimation}
\label{subsection:regularizer}

Many estimation problems, including point-cloud registration, require fitting some model with a set of variable parameters $T$ to a set of measurements. 
Since the parameters $T$ are unknown/poorly known at the start of the estimation, the residuals in $R$ are affected by systematic bias, 
we account for this by adding a secondary regularization term $\beta_j$ to the estimated noise $\sigma_j$ of the system, resulting in Eq.(\ref{eq:InlierRatioReg}).

\begin{equation}
\label{eq:InlierRatioReg}
P(I|R_{i,j}) = \frac{\mathcal{G}(R_{i,j},\alpha_j, \mu_j, \sigma_j + \beta_j)}{ H_j(R_{i,j}) }
\end{equation}

Once the precision of the estimated parameters $T$ improve, the size of $\beta_j$ can be reduced. 
For the point-cloud registration problem we run the registration algorithm until convergence and then reduce $\beta_j$ by 50 percent. Once $\beta_j$ has been decreased, the optimization can be run again. 
We continue this until $\beta_j << \sigma_j$, which means the solution has converged. 

The initial value of $\beta_j$ can be set using prior knowledge about the uncertainty of $T$ or approximated as the standard deviation of the initial set of residuals.

\subsection{Noise normalization}
\label{subsection:noisenormalization}

It is a well known fact that the measurement noise of many sensor
types is correlated to some a priori known factors for the sensor
type. For example, the noise of a measurement by structured light RGBD
cameras is known to increases approximately as the square of the
distance to the sensor.  We formalize this by stating that each
residual value $R_{i,j}$ is either an outlier or sampled with an
individual $\sigma_{i,j} = \sigma_j F(i,j)$. Re-scaling $R_{i,j}$ by
the inverse of $F(i,j)$ makes the scale of the measurement noise
identical for all measurements and the techniques presented in
sections \ref{subsection:trainingstep} and
\ref{subsection:predictionstep} performs as expected. For a residual
computed from the difference between two measurements, the variance of the
resulting residual is the sum of the variances of the two
measurements.

\begin{figure*}[p!]
\begin{tikzpicture}
	\node (tl)  																{ 
		\includegraphics[trim={1 40 3 55},clip, width=0.35\textwidth]{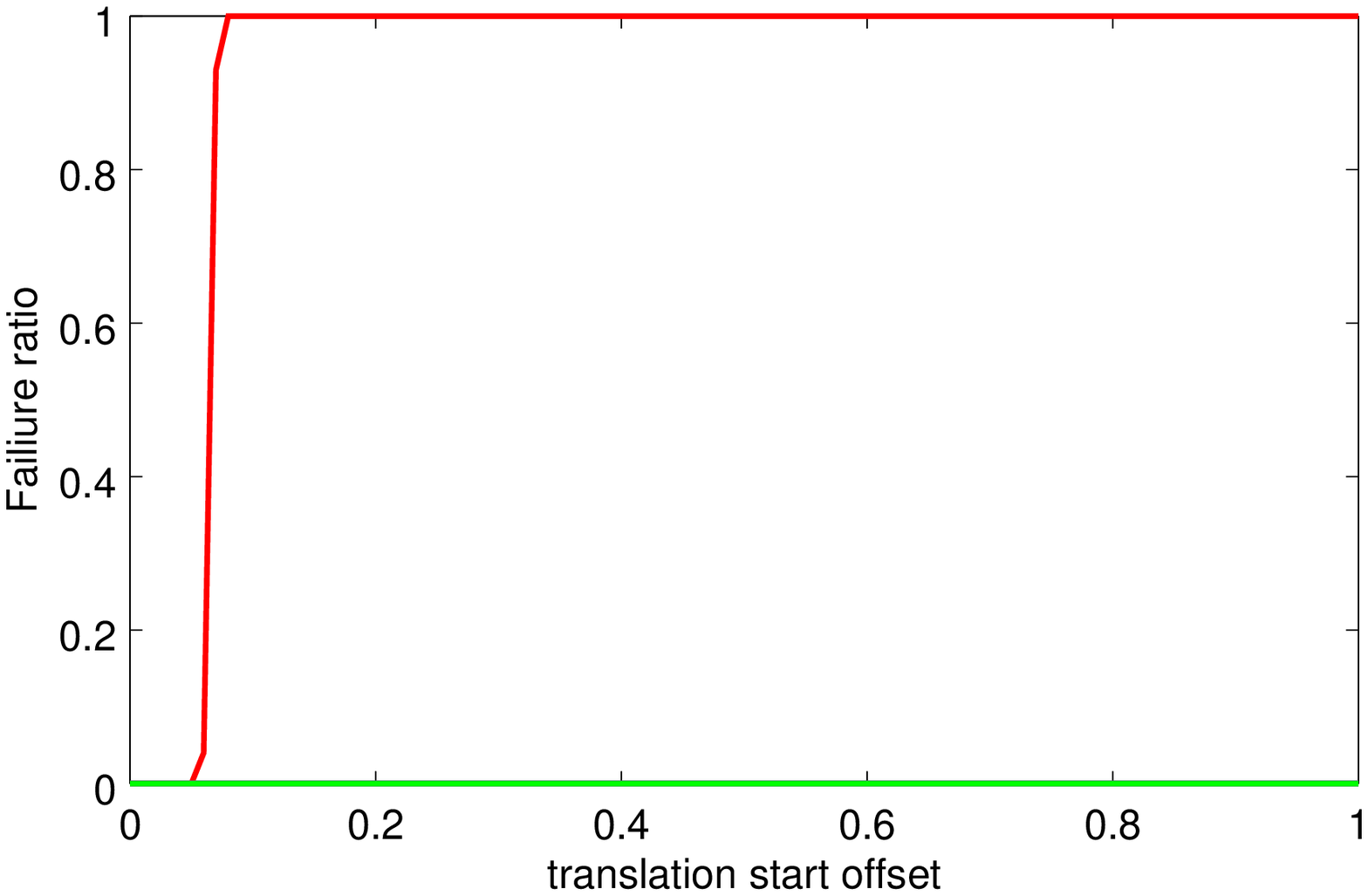}
	};
	\node[right=of tl, node distance=0cm, xshift=-1.7cm,font=\color{red}] (tm) 	{
		\includegraphics[trim={1 40 3 55},clip, width=0.35\textwidth]{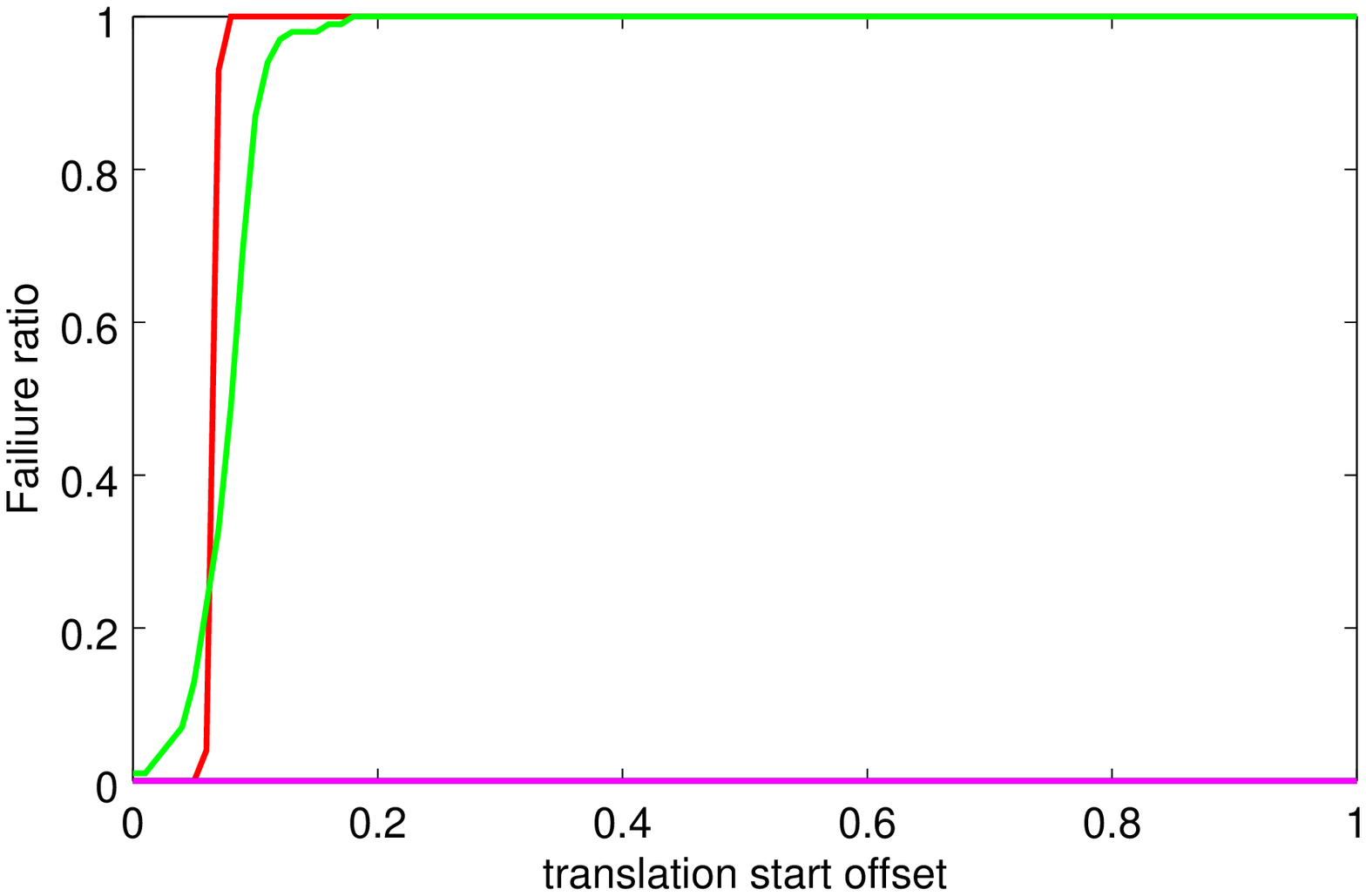}
	};
	\node[right=of tm, node distance=0cm, xshift=-1.7cm,font=\color{red}] (tr)	{
		\includegraphics[trim={1 40 3 55},clip, width=0.35\textwidth]{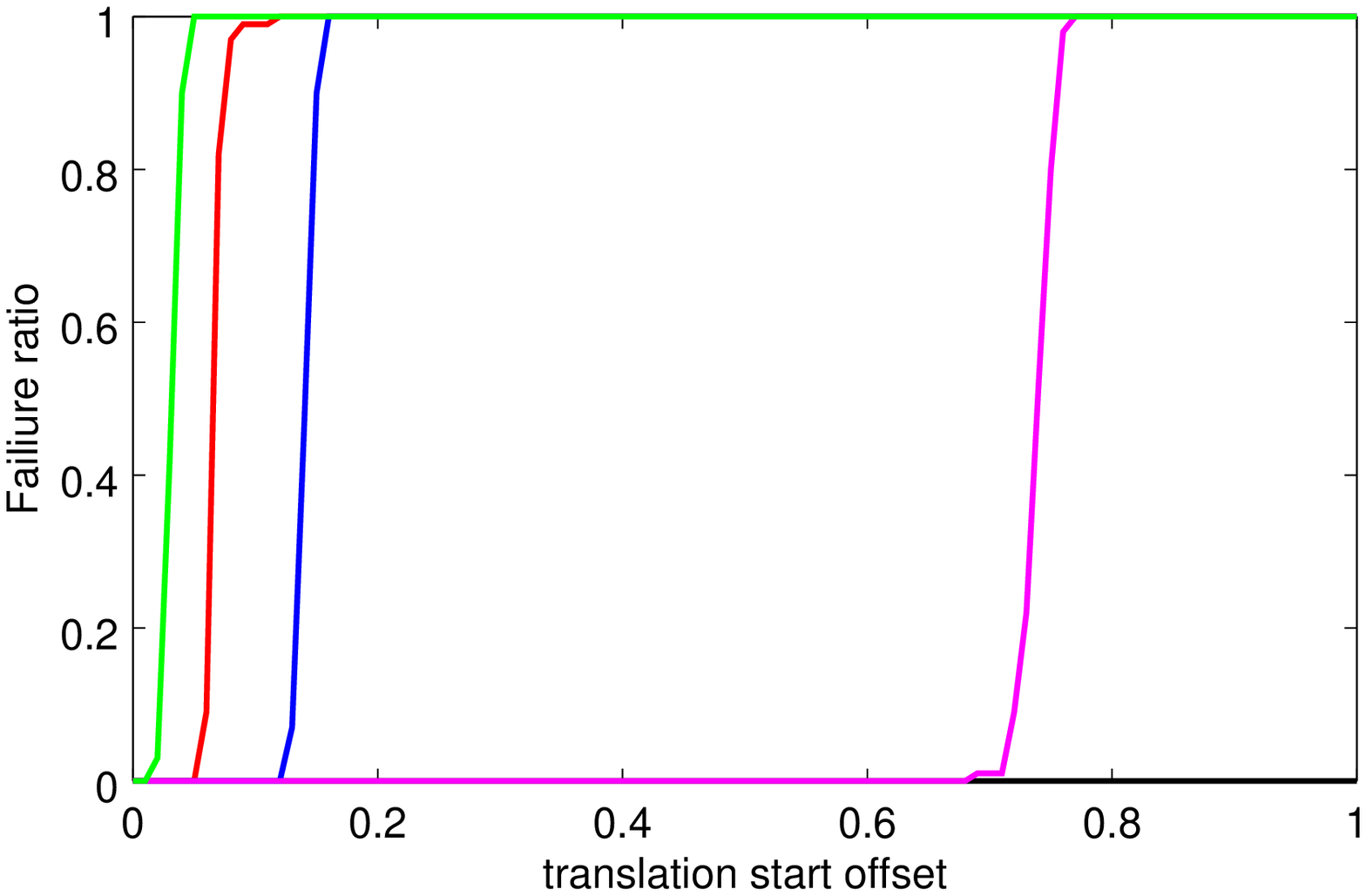}
	};

	\node[below=of tl, node distance=0cm, yshift=1.25cm,font=\color{red}] (bl) {
  		\includegraphics[trim={1 40 3 55},clip, width=0.35\textwidth]{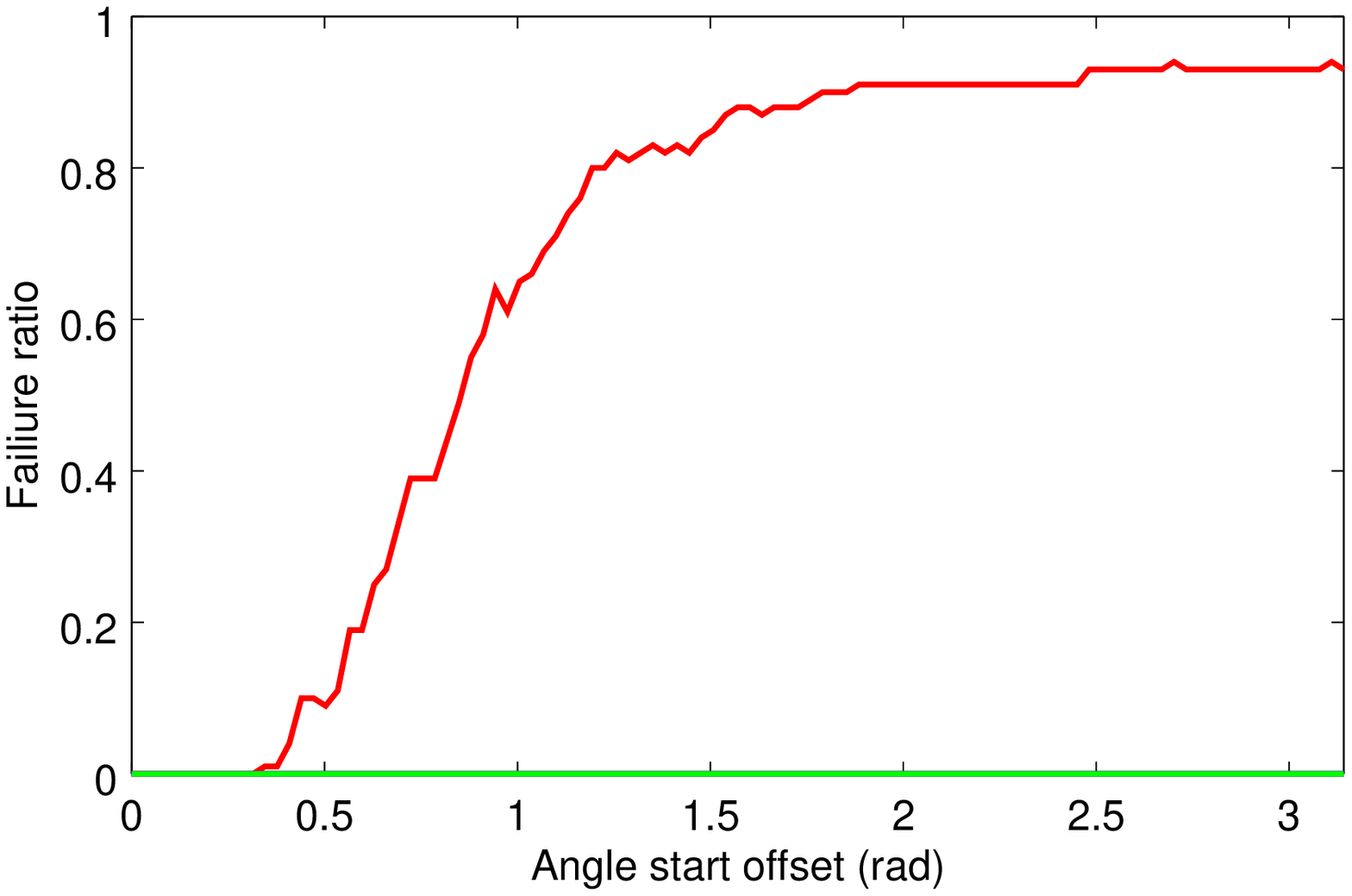}
	};
	\node[right=of bl, node distance=0cm, xshift=-1.7cm,font=\color{red}] (bm) 	{
		\includegraphics[trim={1 40 3 55},clip, width=0.35\textwidth]{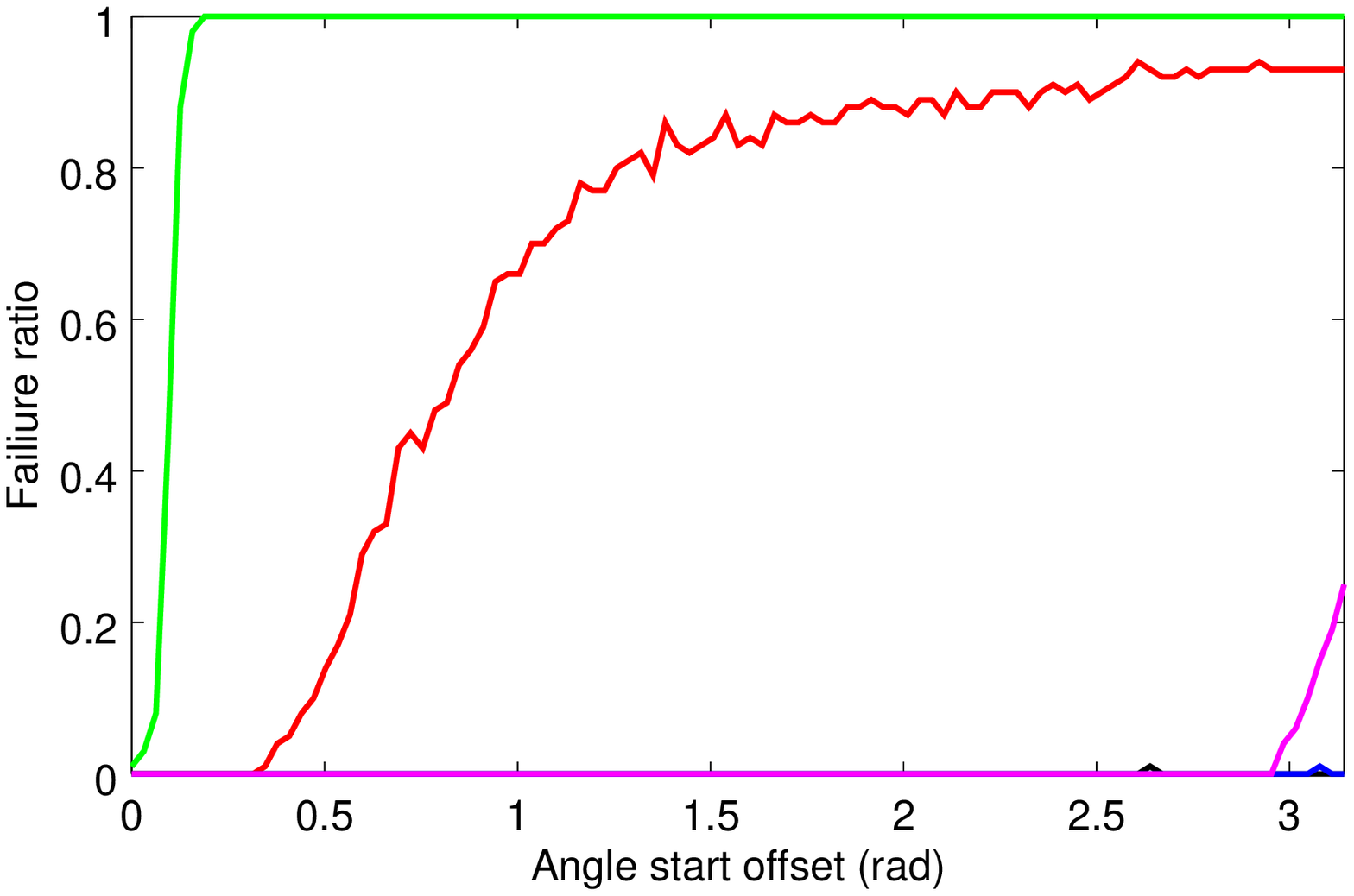}
	};
	\node[right=of bm, node distance=0cm, xshift=-1.7cm,font=\color{red}] (br)	{
		\includegraphics[trim={1 40 3 55},clip, width=0.35\textwidth]{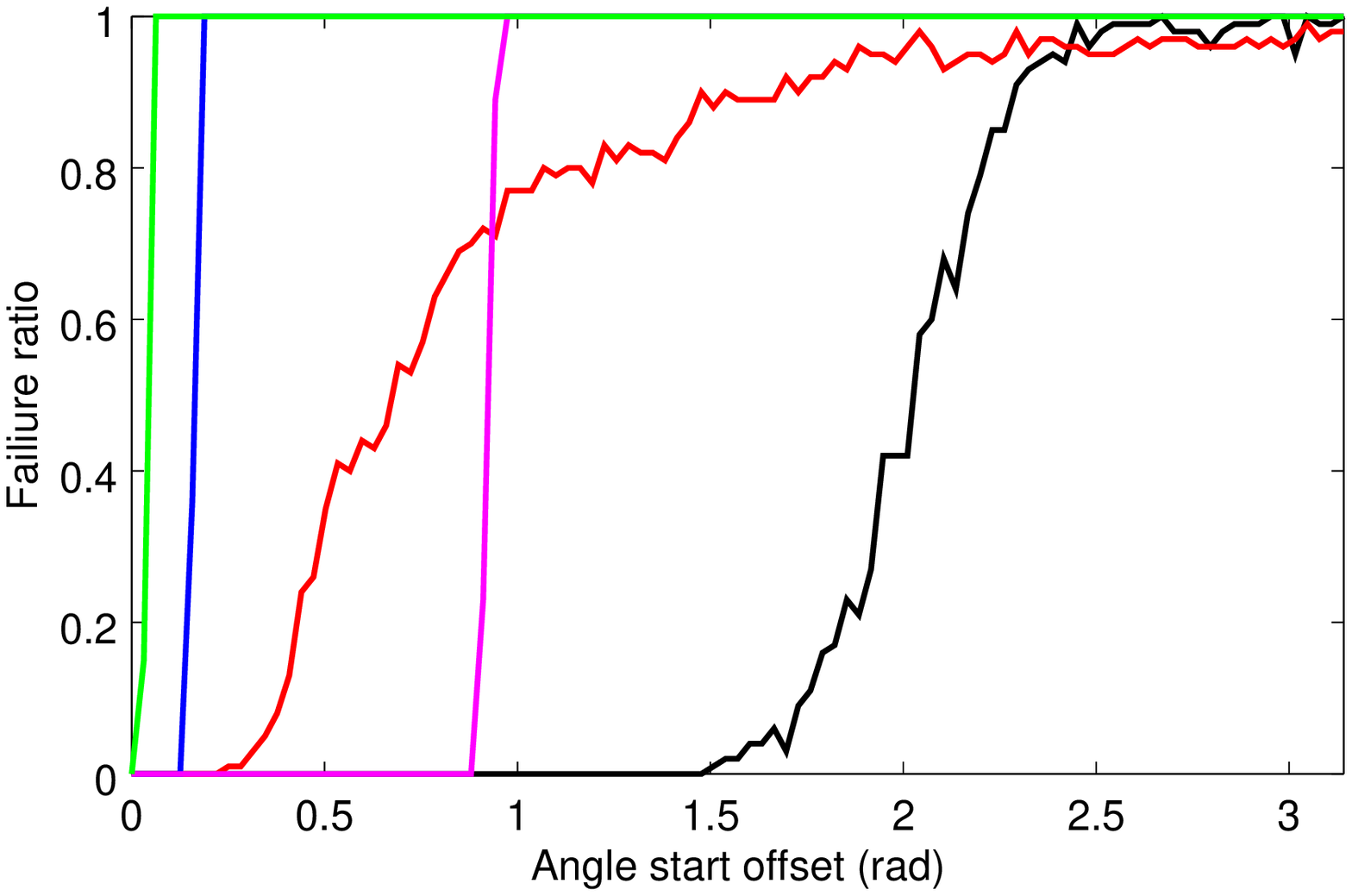}
	};	
  \node[above=of tl, node distance=0cm, yshift=-1.25cm,font=\color{red}]			{1000 Inliers 100 Outliers};
  \node[above=of tm, node distance=0cm, yshift=-1.25cm,font=\color{red}] (test)	{1000 Inliers 1000 Outliers};
  \node[above=of tr, node distance=0cm, yshift=-1.25cm,font=\color{red}] 			{1000 Inliers 10000 Outliers};
  \node[left=of tl, node distance=0cm, rotate=90, anchor=center,yshift=-1.0cm,font=\color{red}] {Translation};
  \node[left=of bl, node distance=0cm, rotate=90, anchor=center,yshift=-1.0cm,font=\color{red}] {Rotation};
  
  \node[above=of test, node distance=0cm, yshift=-1cm,font=\color{black}] (test2) { 
\begin{tikzpicture}
	\draw ( 0.0, 0.5) -- (0,-0.5) -- (17,-0.5) -- (17, 0.5) -- ( 0.0, 0.5);
	\draw[-][draw=black,		very thick] ( 0.0, -0.0)	-- ( 0.0,-0.0) ;
	\draw[-][draw=black,		very thick] ( 0.5, -0.0)	-- ( 1.5,-0.0) node[right] {SIE};
	\draw[-][draw=red,		very thick] ( 2.7, -0.0)	-- ( 3.7,-0.0) node[right] {Truncated L$_2$-norm};
	\draw[-][draw=blue,		very thick] ( 7.2, -0.0)	-- ( 8.2,-0.0) node[right] {L$_1$-norm};
	\draw[-][draw=magenta,	very thick] (10.2, -0.0)	-- (11.2,-0.0) node[right] {L$_{0.1}$-norm};
	\draw[-][draw=green,		very thick] (13.4, -0.0)	-- (14.4,-0.0) node[right] {T-distribution};
\end{tikzpicture}
	};
\end{tikzpicture}
    \caption{Failure ratios for compared solutions with different number of outliers and different numbers of initial transformation estimates. The top row contain experiments of initial transformation translated along the x-axis by 0 to 1 units. The bottom row contain experiments of initial transformation rotated around the x-axis by 0 to $\pi$ radians. In some experiments, some solutions never had any complete failures. This means that the curves overlap of along the x-axis of the figure. Therefore, if a solution is not visible in the figure, no complete failures were recorded. }
	\label{testFail}

\vspace{0.5 cm}
\begin{tikzpicture}
	\node (tl)  																{ 
		\includegraphics[trim={1 40 3 55},clip, width=0.35\textwidth]{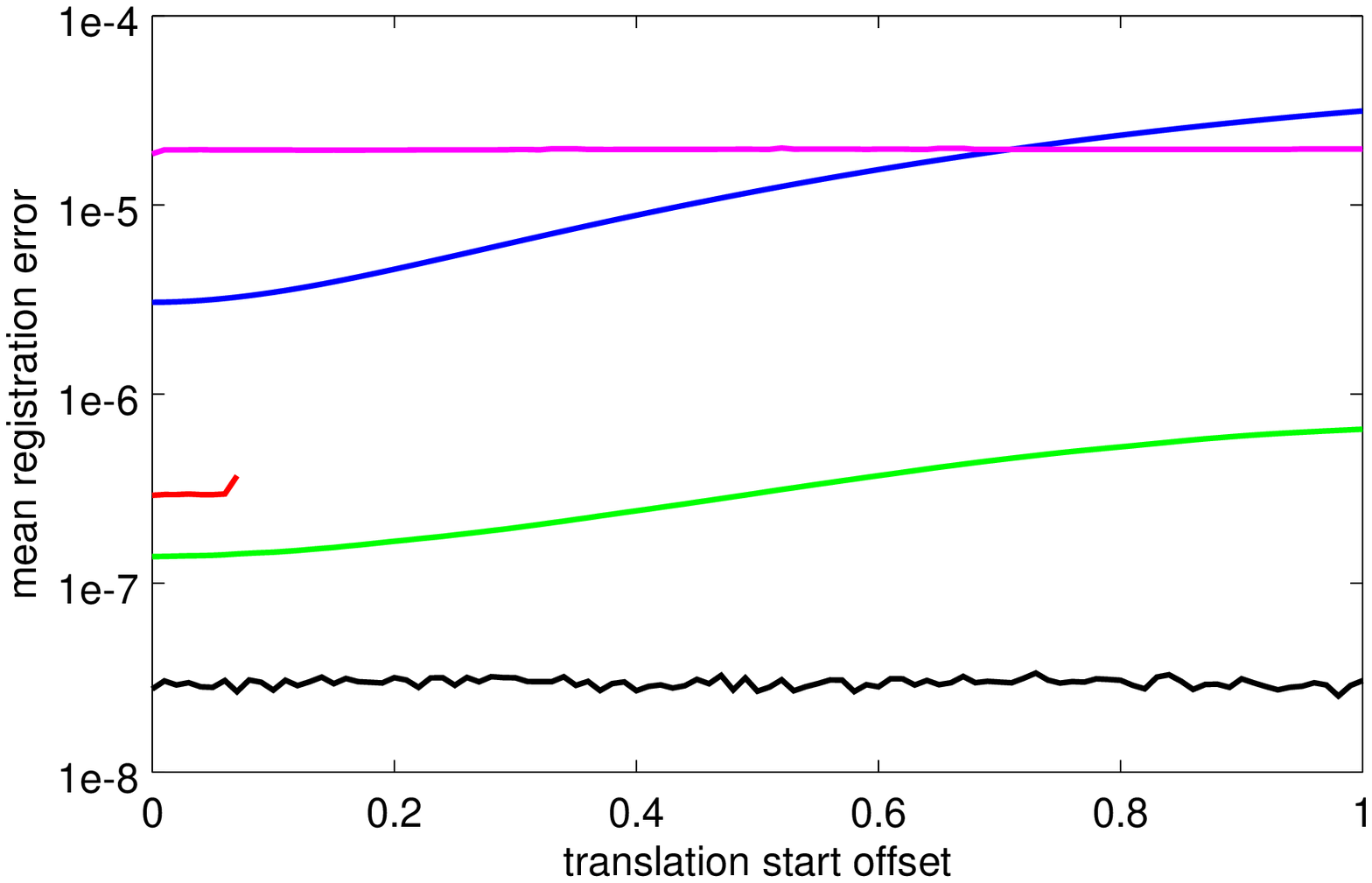}
	};
	\node[right=of tl, node distance=0cm, xshift=-1.7cm,font=\color{red}] (tm) 	{
		\includegraphics[trim={1 40 3 55},clip, width=0.35\textwidth]{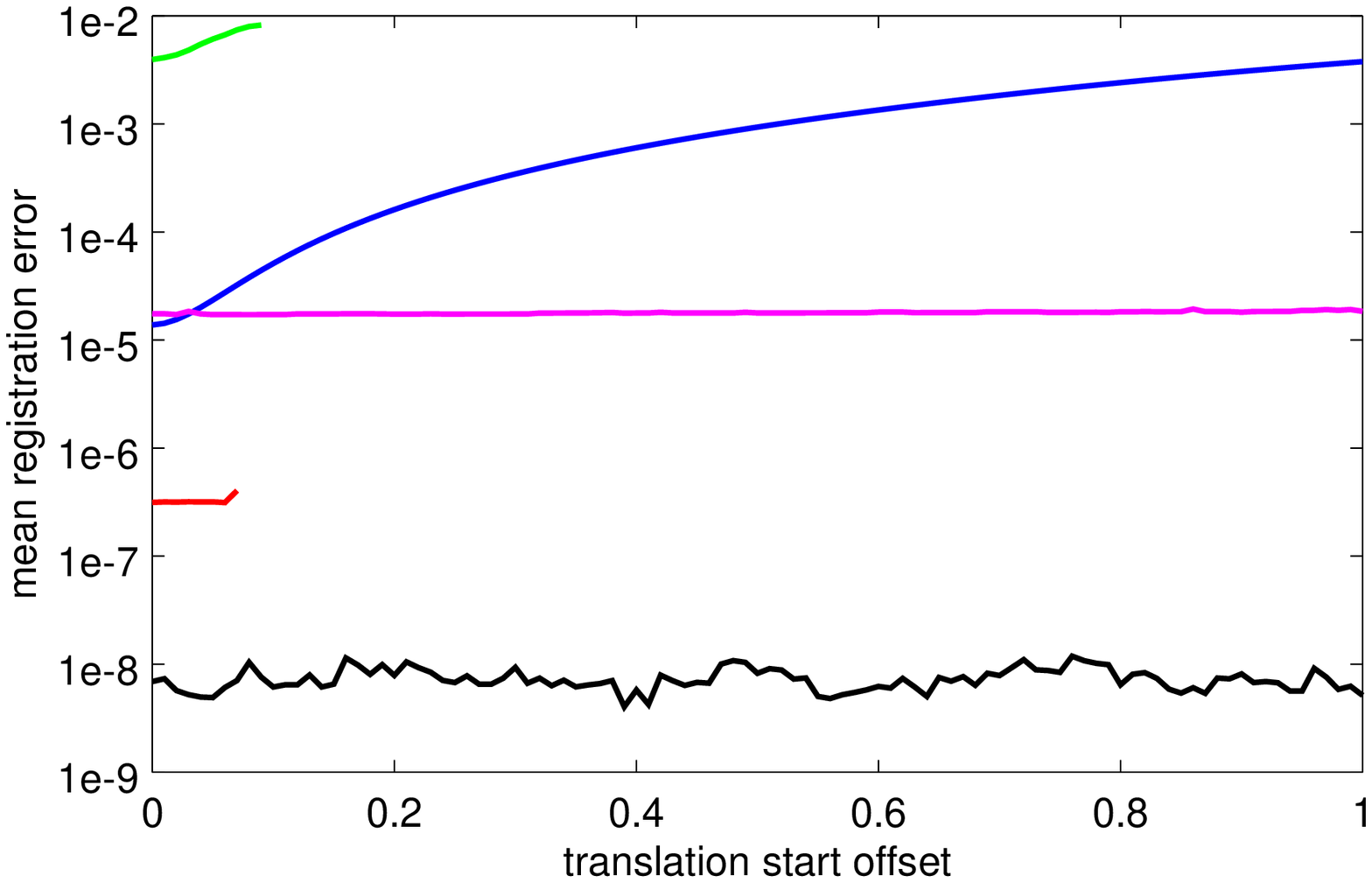}
	};
	\node[right=of tm, node distance=0cm, xshift=-1.7cm,font=\color{red}] (tr)	{
		\includegraphics[trim={1 40 3 55},clip, width=0.35\textwidth]{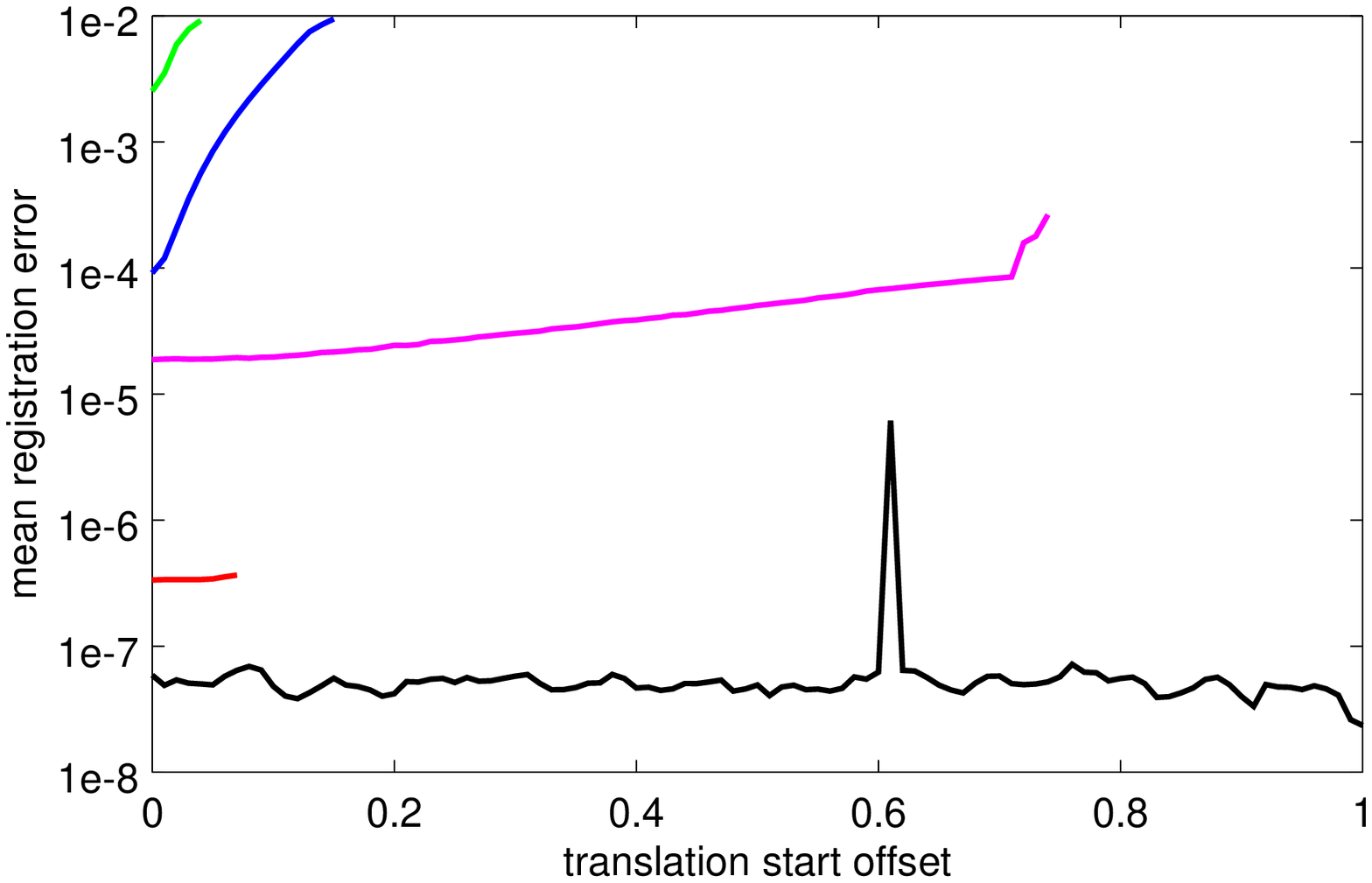}
	};

	\node[below=of tl, node distance=0cm, yshift=1.25cm,font=\color{red}] (bl) {
  		\includegraphics[trim={1 40 3 55},clip, width=0.35\textwidth]{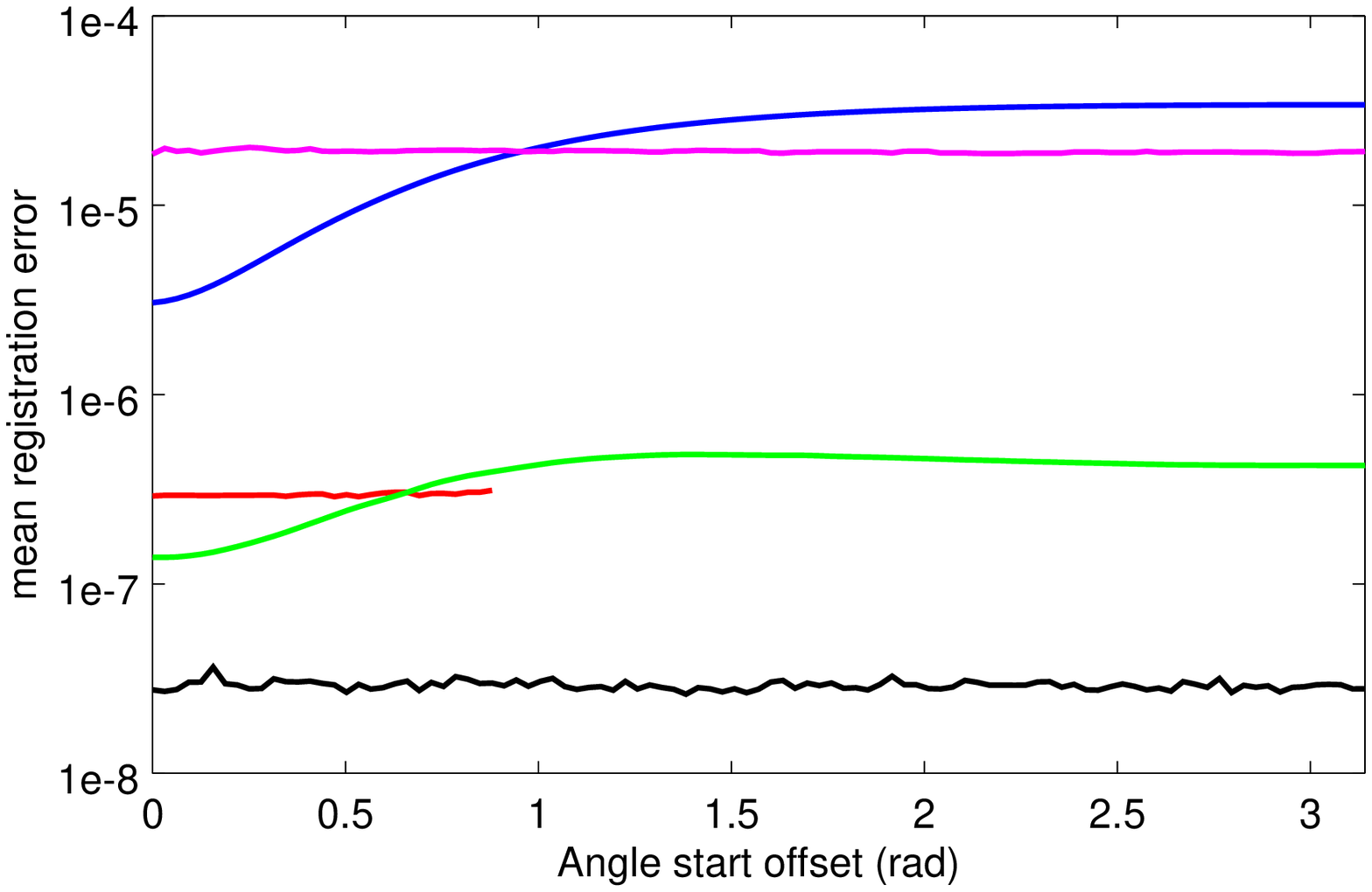}
	};
	\node[right=of bl, node distance=0cm, xshift=-1.7cm,font=\color{red}] (bm) 	{
		\includegraphics[trim={1 40 3 55},clip, width=0.35\textwidth]{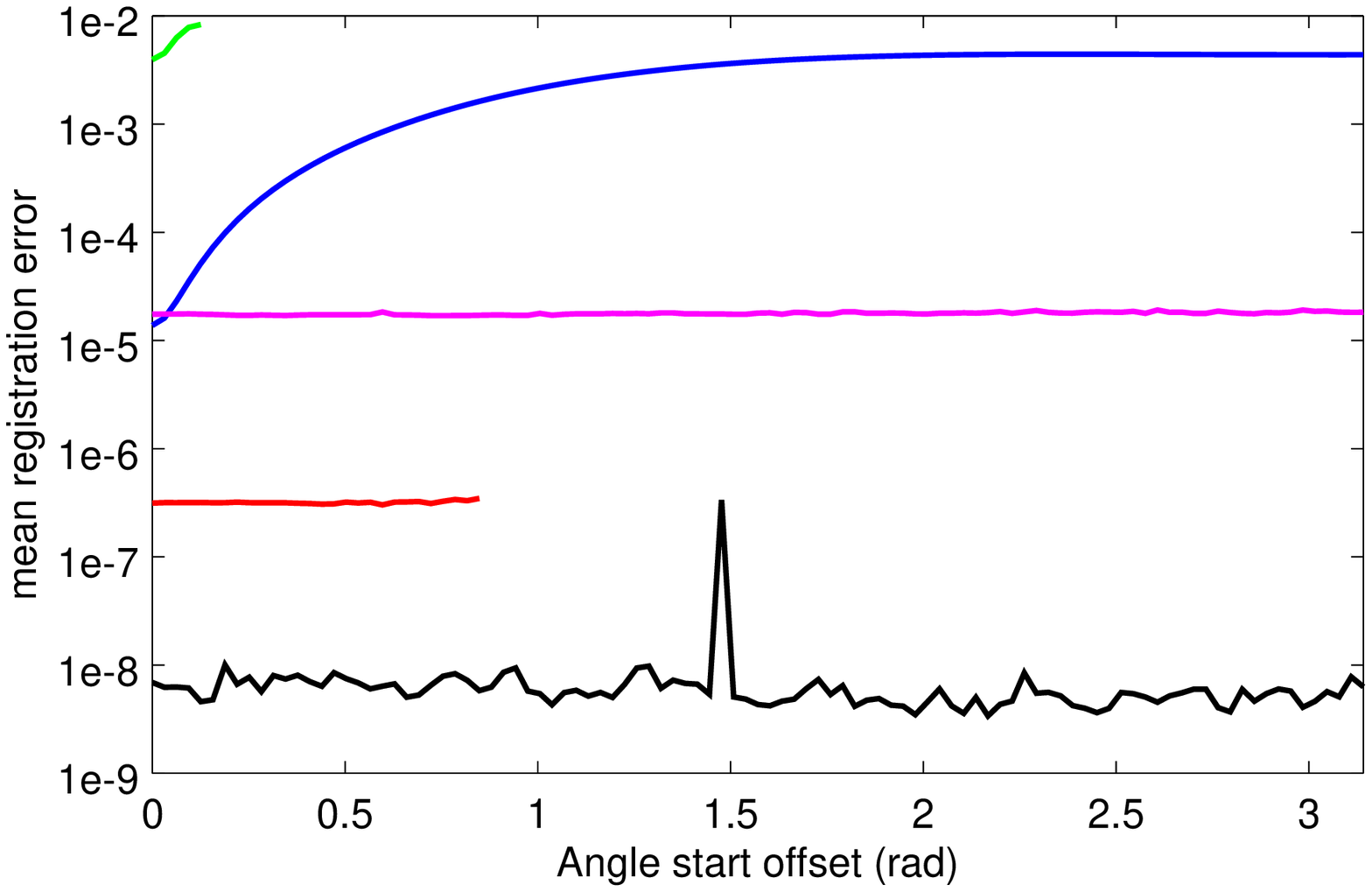}
	};
	\node[right=of bm, node distance=0cm, xshift=-1.7cm,font=\color{red}] (br)	{
		\includegraphics[trim={1 40 3 55},clip, width=0.35\textwidth]{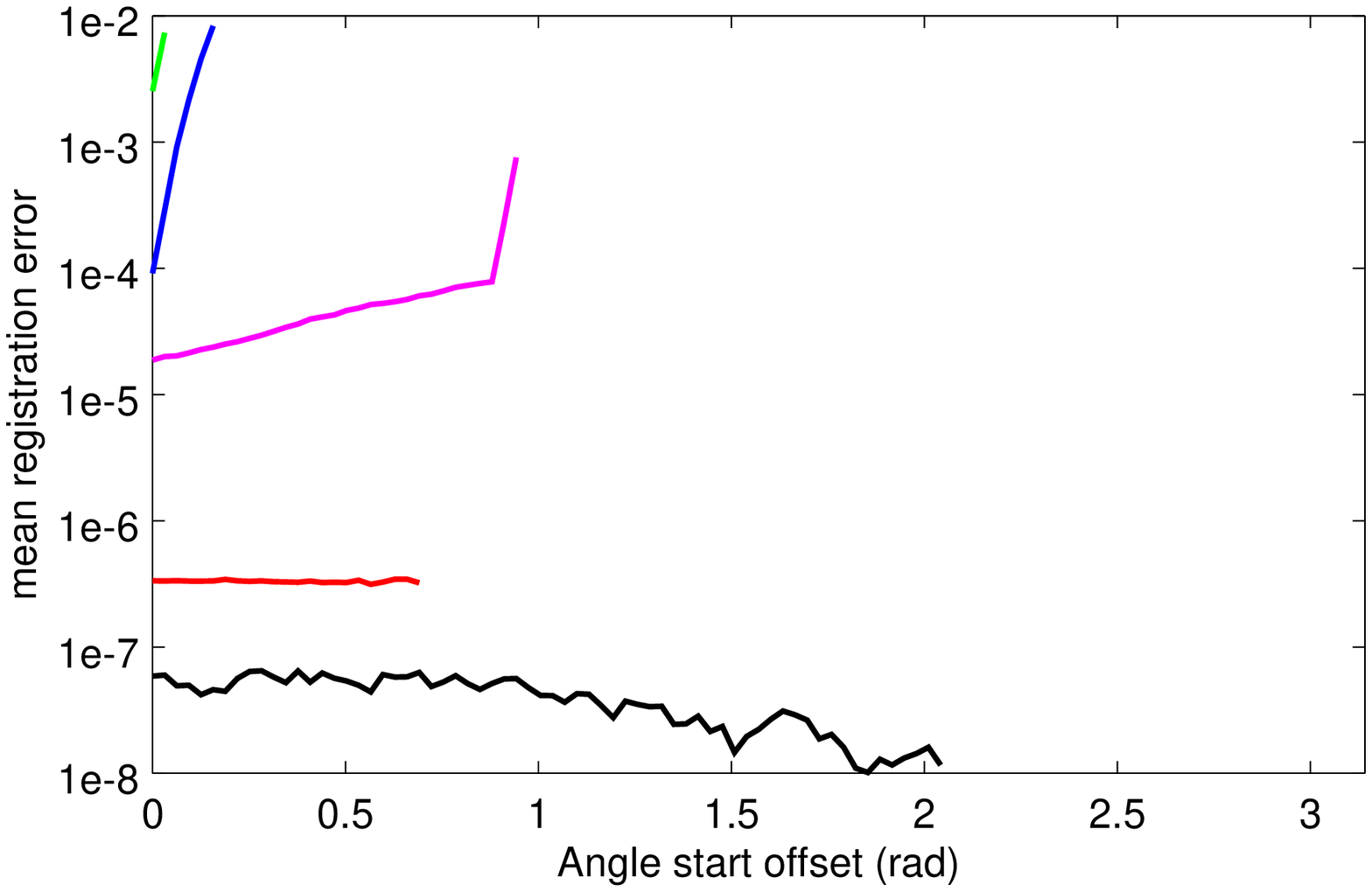}
	};	
  \node[above=of tl, node distance=0cm, yshift=-1.25cm,font=\color{red}]			{1000 Inliers 100 Outliers};
  \node[above=of tm, node distance=0cm, yshift=-1.25cm,font=\color{red}] (test)	{1000 Inliers 1000 Outliers};
  \node[above=of tr, node distance=0cm, yshift=-1.25cm,font=\color{red}] 			{1000 Inliers 10000 Outliers};
  \node[left=of tl, node distance=0cm, rotate=90, anchor=center,yshift=-1.0cm,font=\color{red}] {Translation};
  \node[left=of bl, node distance=0cm, rotate=90, anchor=center,yshift=-1.0cm,font=\color{red}] {Rotation};
  
  \node[above=of test, node distance=0cm, yshift=-1cm,font=\color{black}] (test2) { 
\begin{tikzpicture}
	\draw ( 0.0, 0.5) -- (0,-0.5) -- (17,-0.5) -- (17, 0.5) -- ( 0.0, 0.5);
	\draw[-][draw=black,		very thick] ( 0.0, -0.0)	-- ( 0.0,-0.0) ;
	\draw[-][draw=black,		very thick] ( 0.5, -0.0)	-- ( 1.5,-0.0) node[right] {SIE};
	\draw[-][draw=red,		very thick] ( 2.7, -0.0)	-- ( 3.7,-0.0) node[right] {Truncated L$_2$-norm};
	\draw[-][draw=blue,		very thick] ( 7.2, -0.0)	-- ( 8.2,-0.0) node[right] {L$_1$-norm};
	\draw[-][draw=magenta,	very thick] (10.2, -0.0)	-- (11.2,-0.0) node[right] {L$_{0.1}$-norm};
	\draw[-][draw=green,		very thick] (13.4, -0.0)	-- (14.4,-0.0) node[right] {T-distribution};
\end{tikzpicture}
	};
\end{tikzpicture}

    \caption{ Mean errors of non-failure cases for compared solutions with different number of outliers and different numbers of initial transformation estimates. If the failure rate is greater than 0.5, no mean error is displayed.  The accuracy of estimation vary by orders of magnitudes between the compared solutions, the mean error is therefore drawn on a logarithmic axis. The top row contain experiments of initial transformation translated along the x-axis by 0 to 1 units. The bottom row contain experiments of initial transformation rotated around the x-axis by 0 to $\pi$ radians. }
	\label{testError}
\end{figure*}
\subsection{Adaptive parameters}
\label{subsection:histparam}

Two parameters are required to compute a histogram: the interval in which the histogram is defined and the number of bins in the histogram. Through iterative updating, these values can be computed adaptively to the data.

Given that our choice of $\mathcal{G}_j$ is monotonically decreasing with the distance to the mean value $\mu_j$, one can safely limit $H_j$ to the interval where $\mathcal{G}_j  > \epsilon$, where epsilon is a suitably small number. 

If an initial guess of $\mathcal{G}_j$ is not available, one can safely initialize $\mathcal{G}_j$ to the maximum likelihood estimate assuming that all measurements in $R$ are inliers.

As a means of accounting for sample variance, the number of bins in the histogram can preferably be computed linearly to the number of data that falls within the range of the histogram. This keeps the sample variance of the histogram approximately constant regardless of the amount of data in $H_j$. 

In hopes of further reducing sample variance, the histogram $H_j$ is also smoothed by convolution with a zero mean Gaussian kernel, where the standard deviation is proportional to the width of the histogram. If $\sigma$ is expected to be the same for some set of the dimensions, a joint histogram if all the residuals for those dimensions can be used to reduce the sample variance. Similarly, the absolute values of the residuals can be used to reduce the sample variance of the histograms.

In eq.(\ref{eq:fitParams}) we introduced a penalty term $k$, that bias
the minimization to choose solutions where $H_j(i)\geq
\mathcal{G}(i,\alpha_j,\mu_j,\sigma_j)$. In the ideal case
with no sample variance in $H_j$, one would set $k = \infty$. In
practice, because of sample variance, we fund that $ 1 \leq k \leq 10$
provided good results, with $K = 1$ providing more precise estimations
and $k = 10$ providing more robustness to outliers and poor initial value of $T$. We therefore run the registration with $k = 10$ until convergence and then adapt
the value of $k$ to the data at hand, using the
heuristic $k = P(I | H_j)^{-3}$ where $P(I | H_j)$ is the average probability for the residuals inside the interval of $H_j$ to be an inlier.

During iterative estimation, such as \emph{ICP} minimization, the
parameters for the histogram width, number of bins and value of $k$
can be carried between iterations. This reduces the computational load
of SIE.

\section{EXPERIMENTAL SETUP}
\label{section:EXPERIMENTAL SETUP}

We compare our method to other cost functions used for point-cloud
registration in order to determine both the accuracy and robustness of
the registration relative to other popular solutions. Below we list
the cost functions used during experimentation, and the motivation
behind using them. Many cost functions have names that indicate which
norm is used in the minimization problem. 
The L2-norm is equivalent to assuming that the measurement noise is Gaussian.
All tested solutions are minimized using the iteratively re-weighted least
squares (IRLS) minimization algorithm.

\begin{description}

\item [\textbf{Truncated L2-norm}] \hspace{2.3 cm} The truncated
  L$_2$-norm is a very popular alternative for
  point-cloud-alignment. The idea is to reject all correspondences with
  an error greater than some threshold. The threshold value is usually
  set to some value in the range of three to four $\sigma$. Finding
  the optimal rejection threshold can be hard if one does not know $\sigma$ for the sensor.

\item [\textbf{L$_p$-norm}] \hspace{0.7 cm}
The use of L$_p$-norms, where one artificially set $p$ to a much smaller number than what is actually present in the sensor, is another popular alternative. The idea is that the effect of correspondences with large residuals can be decreased if $p$ is small. A useful property of the algorithm is that knowledge of the sensor noise model is not required, making it a practical solution in many applications where the threshold of the truncated L$_2$-norm is not easily tuned. We will evaluate the result of the L$_1$-norm and L$_{0.1}$-norm as proposed by~\cite{sparseicp_sgp13}.

\item [\textbf{T-distribution}] \hspace{1.7 cm}
In~\cite{kerl13icra} and~\cite{kerl13iros}, the T-distribution is proposed as a self-tuning alternative for registration of RGBD images. Estimation of the scale parameter $\sigma$ is performed using the method from~\cite{kerl13icra}.

\item [\textbf{Statistical Inlier Estimation (ours)}] \hspace{4.7 cm}
SIE models both inlier and outlier distributions. It therefore has the ability to adapt to a wide range of applications. A useful property of the algorithm is that an estimate of the sensor noise is not required, making it a practical solution in many applications where the threshold of the truncated L$_2$-norm is not easily tuned.

\end{description}

\begin{table*}
\caption{Translational RMSE Relative Pose Error(RPE) in m/s for data from~\cite{sturm12iros}.}
\centering
    \begin{tabular}{ | l | l | l | l | l | l | l | p{2.5cm} |}
    \hline
    Sequence 		& Range to noise increase      &  SIE			& Threshold 0.007 m	& L$_1$-norm & L$_{0.1}$-norm	& T-distribution	\\ \hline
    \hline
    \textit{freiburg1 xyz} 	& quadratic			  & 0.023			& \textbf{0.022} 	& 0.038	     & \textbf{0.022	}	& \textbf{0.022}		\\ \hline
    \textit{freiburg1 rpy} 	& quadratic			  & \textbf{0.044}	& 0.045    			& 0.057     	 & 0.045				& 0.049				\\ \hline
    \textit{freiburg1 desk} 	& quadratic			  & \textbf{0.048}	& 0.052				& 0.093     	 & 0.057				& 0.052				\\ \hline
    \textit{freiburg1 desk2}	& quadratic			  & \textbf{0.054}	& 0.065	    			& 0.088     	 & 0.062				& 0.058				\\ \hline
    \textit{freiburg1 room}	& quadratic			  & \textbf{0.062}	& 0.069	    			& 0.095     	 & 0.065				& 0.065				\\ \hline
    \textit{freiburg1 360}	& quadratic			  & \textbf{0.100}	& 0.117 	    			& 0.148     	 & 0.106				& 0.112				\\
    \specialrule{.1em}{.05em}{.05em} 
    \textit{freiburg1 xyz} 	& constant			  & \textbf{0.031}	& 0.037 				& 0.	035      & \textbf{0.031}	& 0.034				\\ \hline
    \textit{freiburg1 rpy} 	& constant			  & \textbf{0.091}	& 0.153    			& 0.132     	 & 0.152				& 0.114				\\ \hline
    \textit{freiburg1 desk} 	& constant			  & \textbf{0.064}	& 0.	093 	    			& 0.100     	 & 0.077				& 0.068				\\ \hline
    \textit{freiburg1 desk2}	& constant			  & \textbf{0.065}	& 0.	091    			& 0.094     	 & 0.078				& 0.071				\\ \hline
    \textit{freiburg1 room}	& constant			  & \textbf{0.081}	& 0.	140    			& 0.129     	 & 0.168				& 0.086				\\ \hline
    \textit{freiburg1 360}	& constant			  & \textbf{0.135}	& 0.296 	    			& 0.200     	 & 0.246				& 0.142				\\ \hline

    \end{tabular}

\label{tumResults}
\end{table*}

\section{SIMULATION}
\label{subsection:SIMULATIONEXPERIMENTS}
As a matter of isolating the effects of the SIE based estimation, we define a simulated experiment where the task is to determine the relative poses of two point-clouds, given a set of point-to-point correspondences, some of which are correct and some which are outliers
\footnote{We define a function $F(k,n,\sigma) = \{A,B\}$ that samples a set of test instance point correspondences $A,B$, where $k$ is the number of inlier matches and $n$ is the number of outlier matches. The $k$ inliers in $A$ are uniformly sampled in $[0,1]^3$ and the inliers in $B$ are identical to the inliers in $A$ but with added Gaussian noise, with a standard deviation of 0.01. The $n$ outliers in $A$ are uniformly sampled in $[0,1]^3$ and each corresponding outlier in $B$ is uniformly sampled around the outlier in $A$ in the interval of $[-1,1]^3$. 

For an initial guess $T_0$, the inlier points are transformed by $T_0$ and the outlier points are kept as is. This ensures that the outlier points do not beneficially change the estimation of $T$.}.

This is equivalent to performing registration using a keypoint matching scheme. If the points in the point-clouds are independently sampled with zero mean Gaussian noise and transformed by some rigid body transform $T$, 
the maximum likelihood estimate $T_{MLE}$ of $T$ is given the least squares fit of the correct correspondences.

Our system aims to compute an estimate of $T$, which is as close to $T_{MLE}$ as possible. 
In this paper, we measure the Accuracy of estimation, robustness to outliers and robustness to initial guess of $T$.
Since this data is simulated, we can vary the number of outliers and change the initial guess for the transformation.

The test error of an estimated transform is defined as the difference between the root-mean-square-error (RMS) of 
the correct correspondences and the corresponding value for $T_{MLE}$. 
The advantage of this error metric is that we are directly measuring how much less probable the estimated solution is to be correct, than the best possible solution. We consider a tested transform with an error greater than the measurement noise a complete failure.

To acquire reliable statistics we sample a set of 100 different test instances on which we apply our test initial guesses. A total score for an initial transformation and a set of outliers is then computed as the mean error for all the non-failure tests. This score is useful to determine the precision of a registration algorithm. Looking at the ratio of failures is a good method for determining the robustness of the estimator used.

\subsection{Simulation Results}
\label{subsection:simulationsetup}
We test three different cases, an easy case with 1000 inliers and 100 outliers, a medium case with 1000 inliers and 1000 outliers, and a hard case with 1000 inliers and 10000 outliers. 
To test the range of convergence, two sets of test transformations are created, one where translation on the X-axis in the range of $[0,1]$ is tested and the other where rotation around the X-axis is tested in the range of $[0,\pi]$ radians. 
From figures \ref{testFail} and \ref{testError}, we observe that SIE reliably and significantly outperform the other cost functions with regards to robustness to poor initialization for $T_0$, robustness to outliers and precision of estimation.


\section{PRIMESENSE DATA}
In ~\cite{sturm12iros}, a benchmark for RGBD-image slam and visual odometry was presented. ~\cite{sturm12iros} defines two measurement errors: relative-pose-error (RPE) and absolute-trajectory-error (ATE). The translation RPE is used to test visual odometry and registration algorithms, making it suitable for our purposes.

As previously stated, the noise of a measurement $i$ in the primesense cameras increase approximately by the distance to the sensor squared. We therefore try both $\sigma_i = z_i^2$ and $\sigma_i = 1$ and scale the residuals accordingly for all tested methods. From experience, we know that the noise of the primesense cameras is in the range of 0.001 to 0.002 m. We set the threshold for the truncated L$_2$-norm is set to 0.007 m. For the other algorithms all parameters are kept identical to the simulation experiments, despite there being a difference of an order of magnitude in the size of the noise.

For computational reasons, the registration algorithm is limited to 2000 randomly sampled nearest neighbour matches for all cost functions. We replace the point-to-point distance used in section~\ref{subsection:SIMULATIONEXPERIMENTS} with the point-to-plane distance metric.

From the results in table~\ref{tumResults}, we can clearly see that SIE consistently outperforms the other cost functions, SIE is has the lowest score or shared lowest score in 11 out of 12 tests. SIE does well especially on the hard cases where all the costfunctions have a high error score. It is also clear that knowing the connection between range and measurement noise has a clear positive effect on the estimation for the tested solutions.


\section{CONCLUSIONS}

In this paper we introduced a statistical inlier estimation (SIE) 
system. SIE is used to determine the probability of pairwise
matches to be correct without the need hand tuned parameters. This
makes using the system simple and flexible. SIE is intrinsically
portable and we have tried it in two different scenarios for sensors with vastly different amounts of noise, 
without changing any parameters. We extend the \emph{ICP} algorithm to utilize the SIE
system for outlier rejection and show that the improved \emph{ICP} algorithm outperforms
comparable cost functions on the dataset
from~\cite{sturm12iros}. Similarly, we show on synthetic data that SIE improves both accuracy and robustness when
estimating transforms from point-to-point correspondences,
especially in the case of many outliers and poor initialization
values.


\appendix
\label{appendix}
In the main body of this paper, we made the assumption that the sensor measurement noise was Gaussian. This is not required by the SIE system presented in section~\ref{section:inlierestimation}. If we simulate a set of inliers similarly to section~\ref{section:EXPERIMENTAL SETUP}, but with Laplacian noise instead of Gaussian, we can investigate if SIE can be used to estimate the parameters of the Laplacian noise. Both the Laplacian and Gaussian distributions are subsets of the Generalized Gaussian Distribution (GGD)

\begin{equation}
\label{eq:point2surfaceprob}
\mathcal{G}(x,\mu,\sigma,p) = \frac{p}{2 \sigma \Gamma ( p^{-1} )} e^{- (\frac{| x-\mu|}{\sigma})^p}
\end{equation}
where $\Gamma$ denotes the gamma function. The Laplacian distribition is the special case of a GGD when $p = 1$ and the Gaussian distribution the special case of a GGD when $p = 2$.

The equivalent of eq.(\ref{eq:regprob}, with Gaussian noise becomes 

\begin{equation}
\label{eq:regprobGGD}
\begin{split}
& \argmin_{b,T} -log \Big( P(A|b,T) \Big) = \argmin_{b,T} \sum_{i=0}^{n} \Big( \frac{||r_i||}{\sigma_i} \Big)^{p_i}
\end{split}
\end{equation}
which is equivalent to minimizing over the L$_p$-norm instead of the L$_2$-norm. The minimization using the probabilities from SIE can be performed using the IRLS algorithm where 

\begin{equation}
\label{eq:regprob3GGD}
\begin{split}
w(r_i) = \sigma_i^{-p_i} max(||r_i||,\delta)^{(p_i-2)} P(I|r_i) 
\end{split}
\end{equation}
where $\delta$ is a very small number which makes sure that division by zero is avoided. 

For the Laplacian distribution, the test error of an estimated transform is defined as the difference between the mean-absolute-error of 
the correct correspondences and the corresponding value for $T_{MLE}$. Similarly to the RMS error for the Gaussian distribution, the advantage of the mean-absolute-error metric for Laplacian data is that we directly measure the probability of the estimated solution.

\begin{figure*}
\begin{tikzpicture}
	\node (tl)  																{ 
		\includegraphics[trim={1 40 3 55},clip, width=0.35\textwidth]{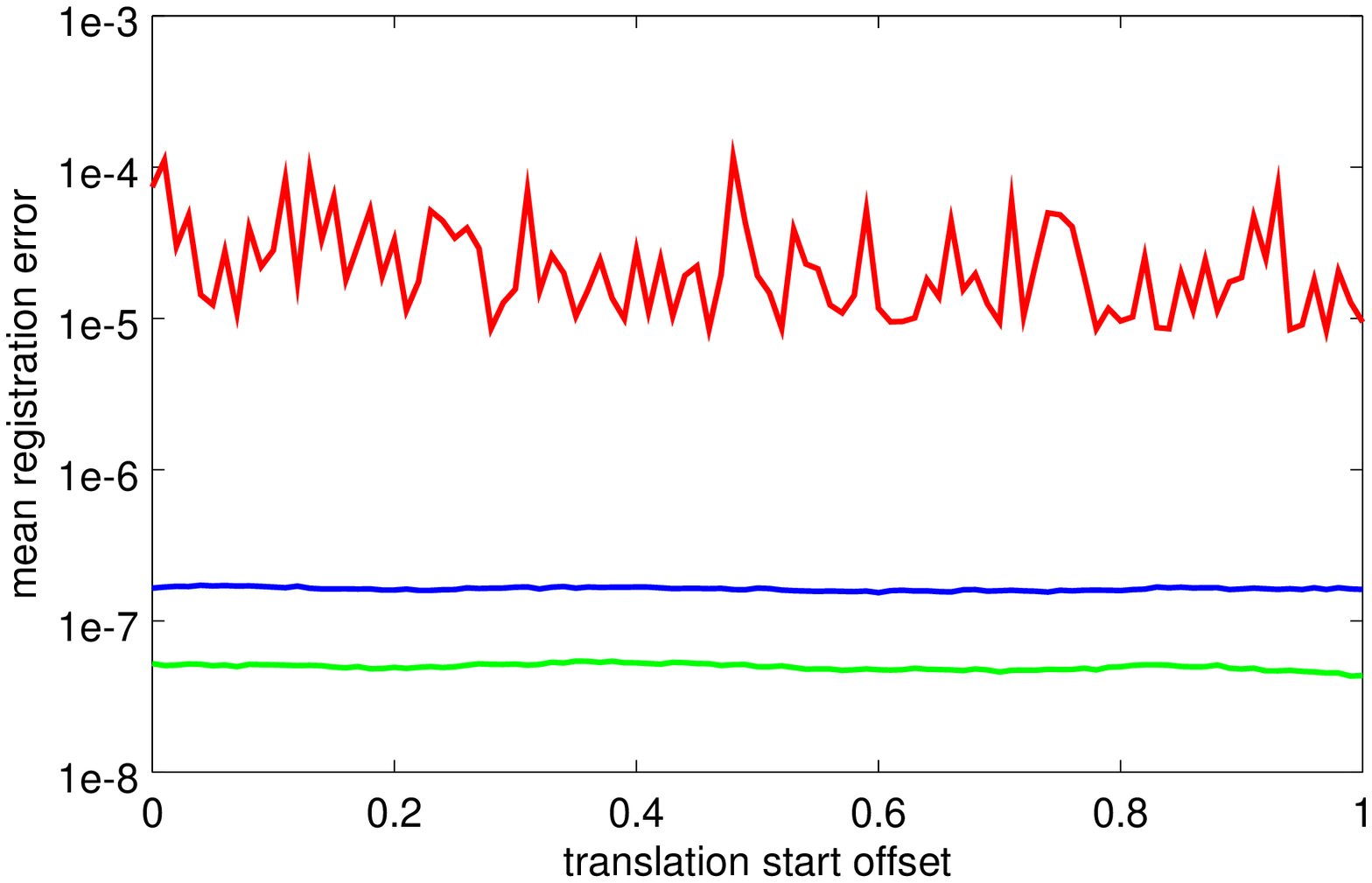}
	};
	\node[right=of tl, node distance=0cm, xshift=-1.7cm,font=\color{red}] (tm) 	{
		\includegraphics[trim={1 40 3 55},clip, width=0.35\textwidth]{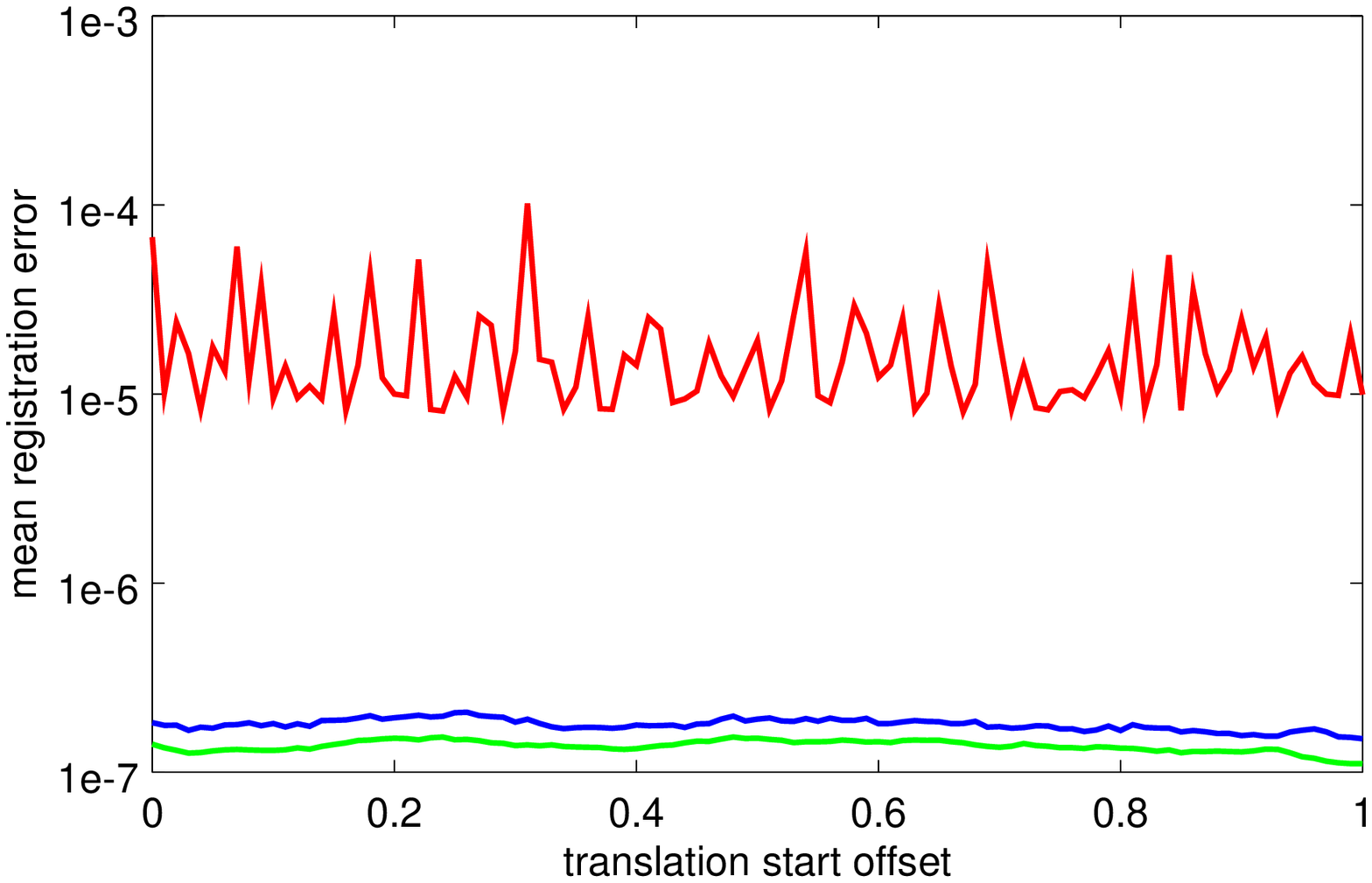}
	};
	\node[right=of tm, node distance=0cm, xshift=-1.7cm,font=\color{red}] (tr)	{
		\includegraphics[trim={1 40 3 55},clip, width=0.35\textwidth]{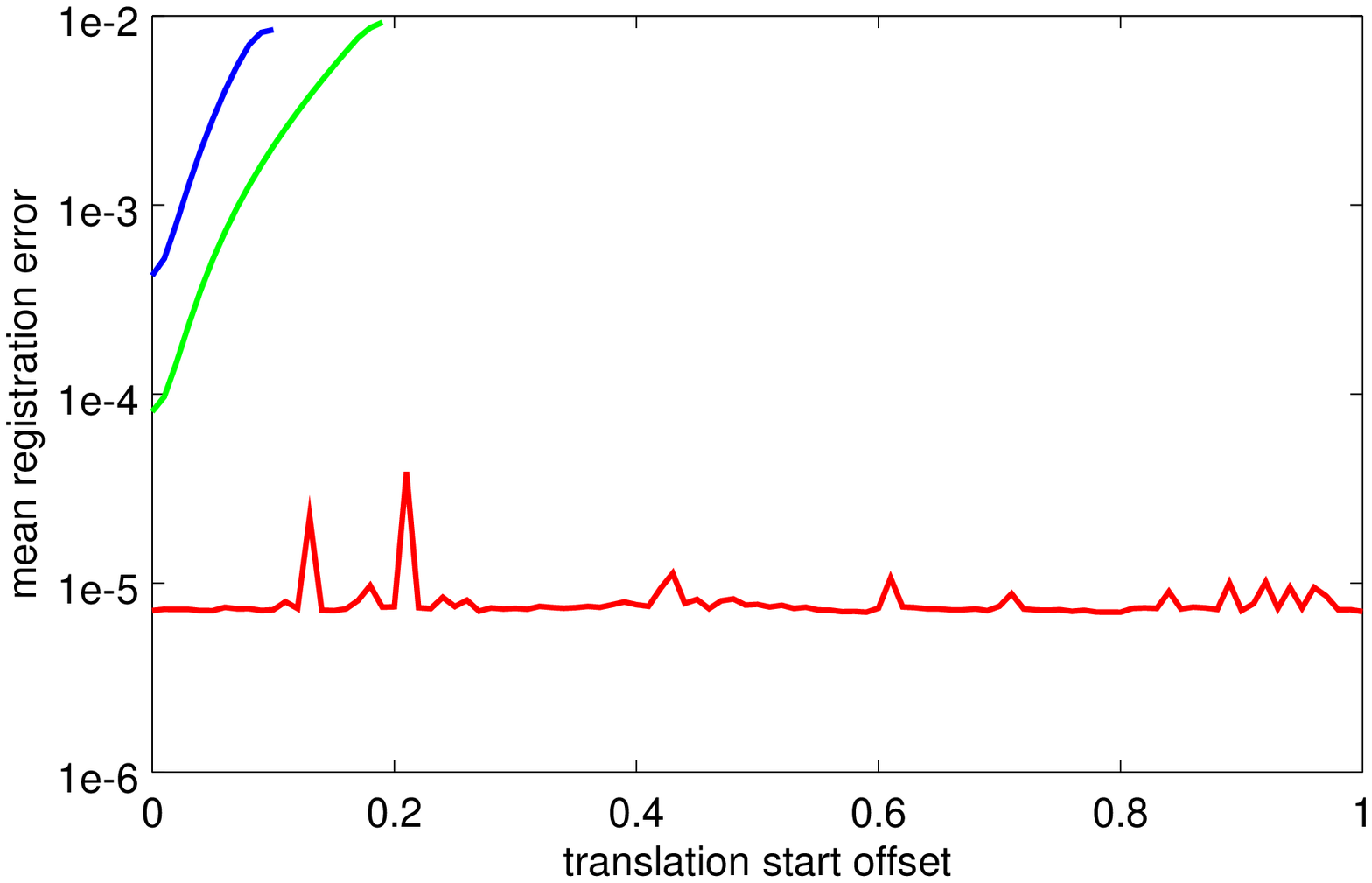}
	};

	\node[below=of tl, node distance=0cm, yshift=1.25cm,font=\color{red}] (bl) {
  		\includegraphics[trim={1 40 3 55},clip, width=0.35\textwidth]{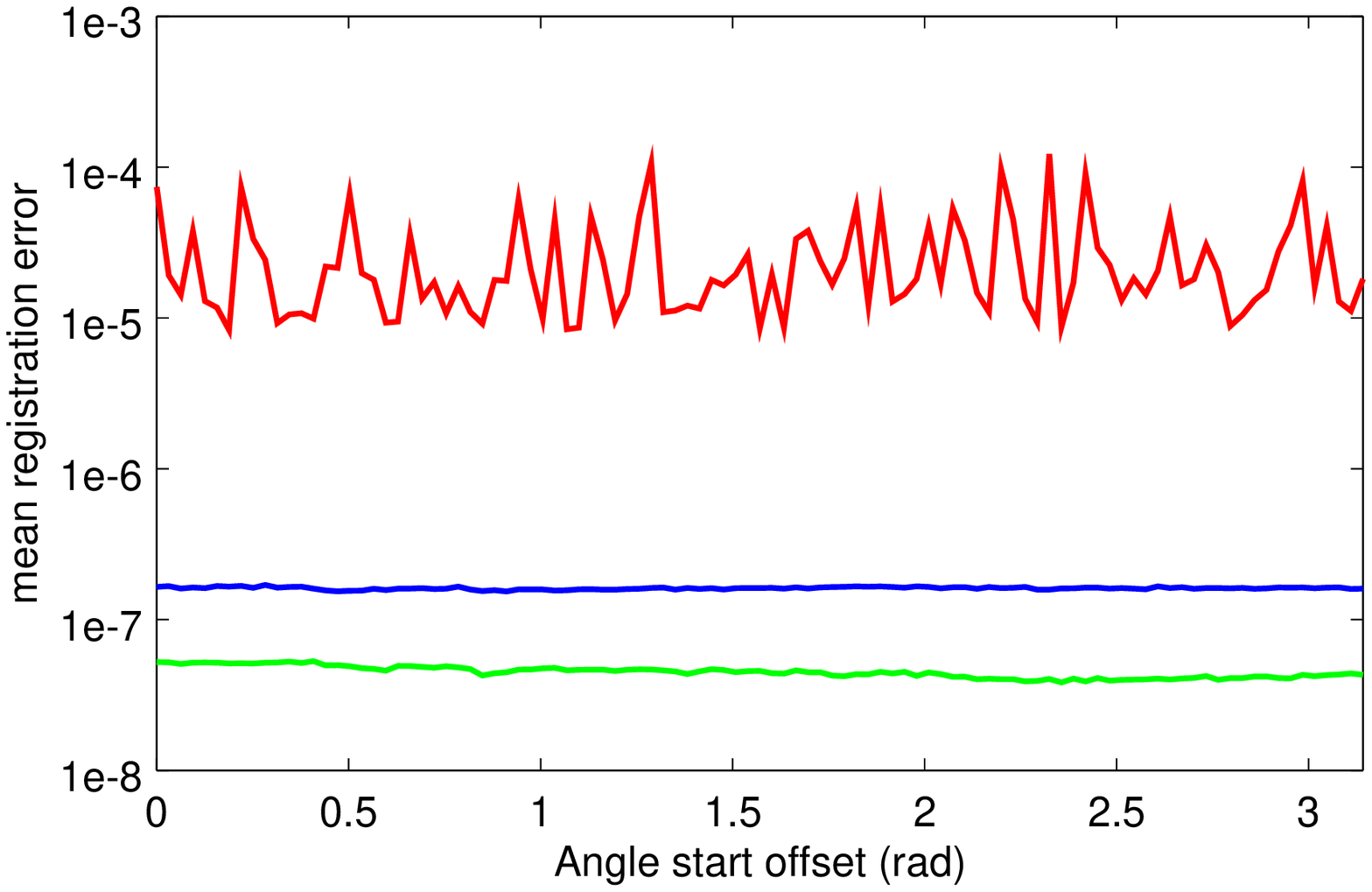}
	};
	\node[right=of bl, node distance=0cm, xshift=-1.7cm,font=\color{red}] (bm) 	{
		\includegraphics[trim={1 40 3 55},clip, width=0.35\textwidth]{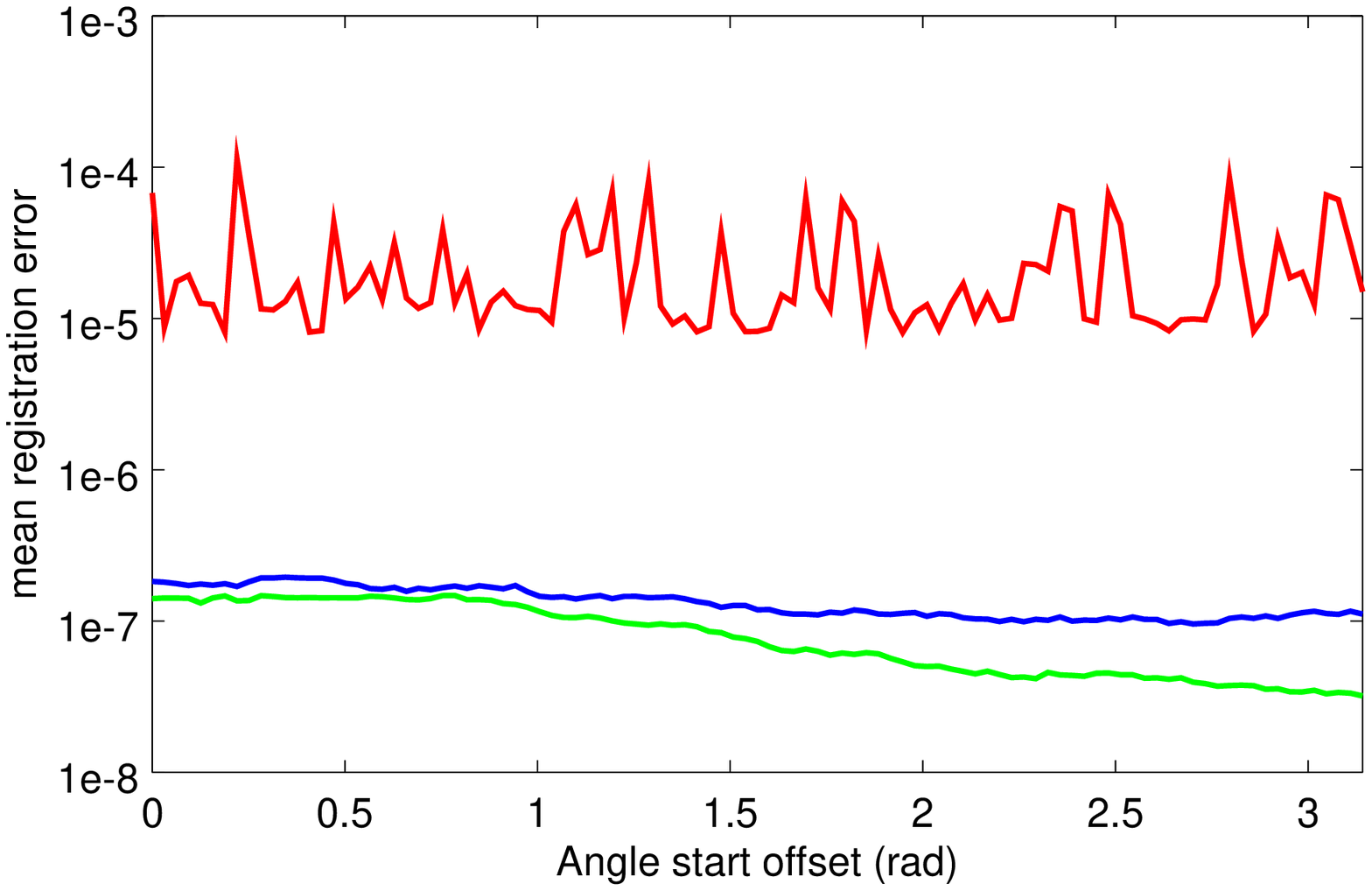}
	};
	\node[right=of bm, node distance=0cm, xshift=-1.7cm,font=\color{red}] (br)	{
		\includegraphics[trim={1 40 3 55},clip, width=0.35\textwidth]{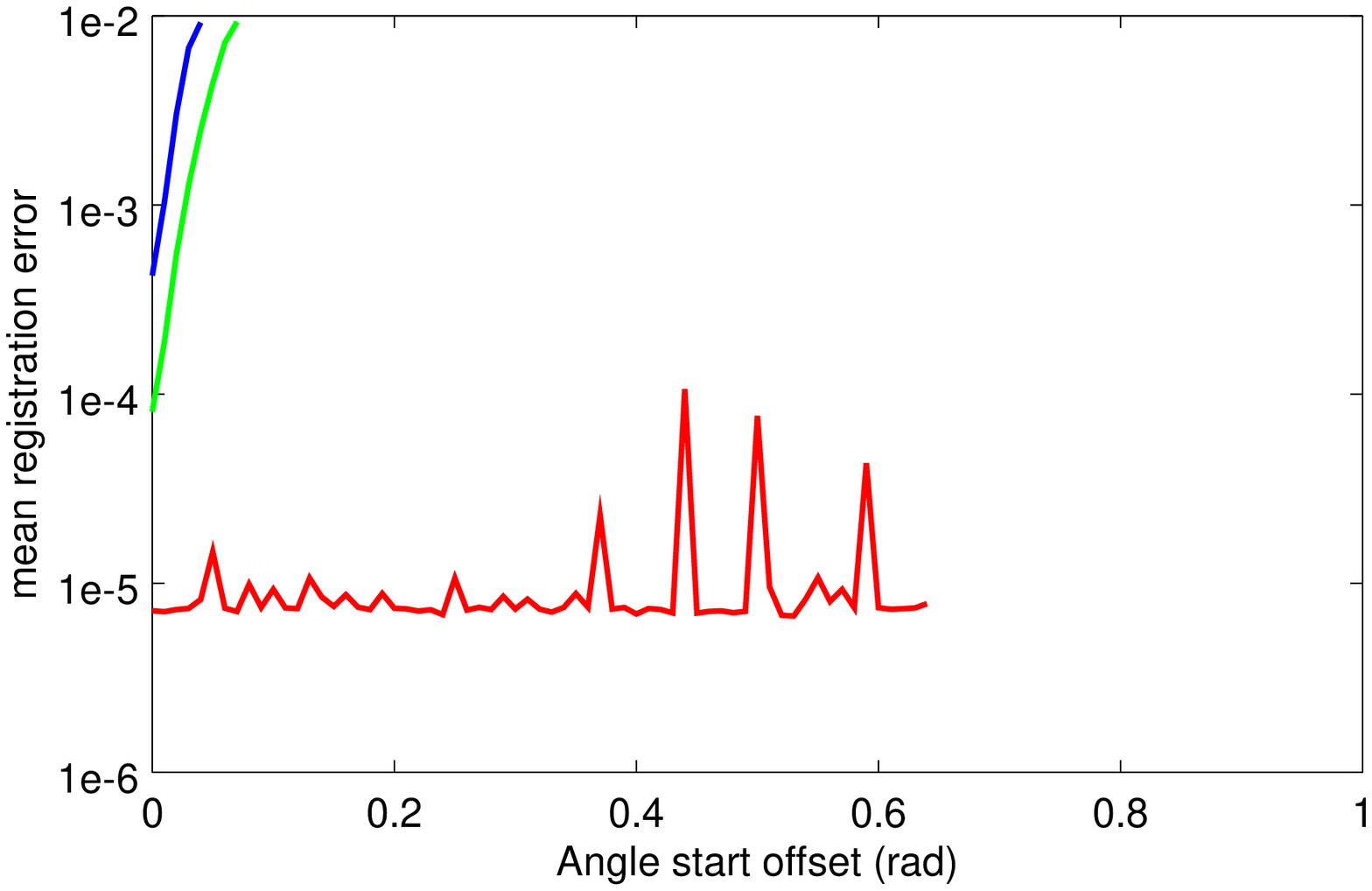}
	};	

  \node[above=of tm, node distance=0cm, yshift=0.5cm,font=\color{black}]			{\huge Laplacian noise};
  \node[above=of tl, node distance=0cm, yshift=-1.25cm,font=\color{red}]			{1000 Inliers 100 Outliers};
  \node[above=of tm, node distance=0cm, yshift=-1.25cm,font=\color{red}] (test)	{1000 Inliers 1000 Outliers};
  \node[above=of tr, node distance=0cm, yshift=-1.25cm,font=\color{red}] 		{1000 Inliers 10000 Outliers};
  \node[left=of tl, node distance=0cm, rotate=90, anchor=center,yshift=-1.0cm,font=\color{red}] {Translation};
  \node[left=of bl, node distance=0cm, rotate=90, anchor=center,yshift=-1.0cm,font=\color{red}] {Rotation};
  
  \node[above=of test, node distance=0cm, yshift=-1cm,font=\color{black}] (test2) { 
\begin{tikzpicture}
	\draw ( 0.0, 0.5) -- (0,-0.5) -- (17,-0.5) -- (17, 0.5) -- ( 0.0, 0.5);
	\draw[-][draw=black,		very thick] ( 0.0, 0)	-- ( 0.0,0) ;
	\draw[-][draw=red,		very thick] ( 1.0, 0)	-- ( 3.0,0) node[right] {  SIE L$_2$-norm};
	\draw[-][draw=green,		very thick] ( 6.0, 0)	-- ( 8.0,0) node[right] {  SIE L$_1$-norm};
	\draw[-][draw=blue,		very thick] ( 11.0, 0)	-- ( 13.0,0) node[right] {  SIE estimated norm};
\end{tikzpicture}
	};  
  
\end{tikzpicture}
    \caption{ Mean errors of non-failure cases for compared solutions with different number of outliers and different numbers of initial transformation estimates. The measurements are sampled with Laplacian noise. If the failure rate is greater than 0.5, no mean error is displayed.  The accuracy of estimation vary by orders of magnitudes between the compared solutions, the mean error is therefore drawn on a logarithmic axis. The top row contain experiments of initial transformation translated along the x-axis by 0 to 1 units. The bottom row contain experiments of initial transformation rotated around the x-axis by 0 to $\pi$ radians. }
	\label{testErrorLaplacian}
\vspace{0.5 cm}
\begin{tikzpicture}
	\node (tl)  																{ 
		\includegraphics[trim={1 40 3 55},clip, width=0.35\textwidth]{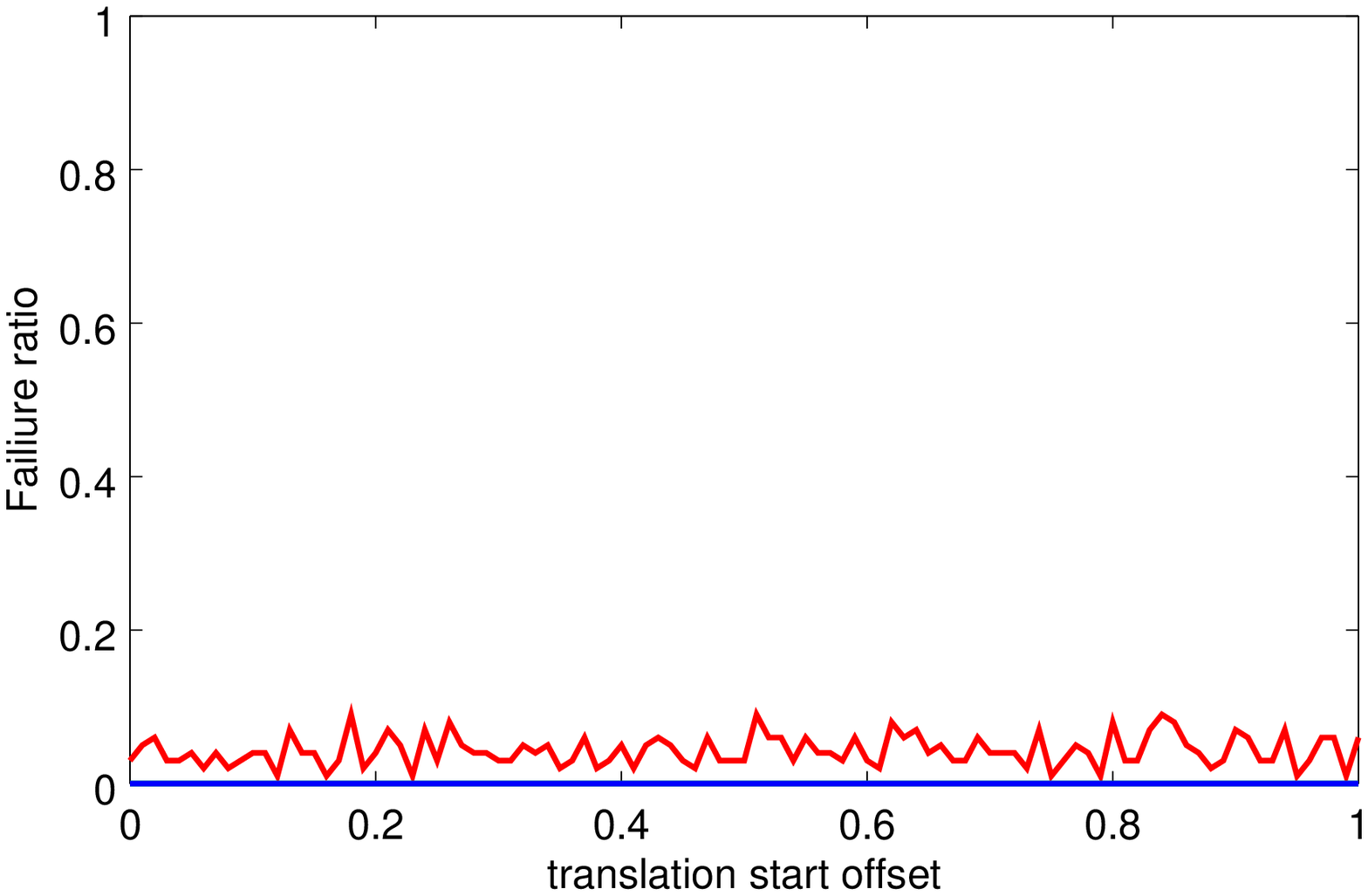}
	};
	\node[right=of tl, node distance=0cm, xshift=-1.7cm,font=\color{red}] (tm) 	{
		\includegraphics[trim={1 40 3 55},clip, width=0.35\textwidth]{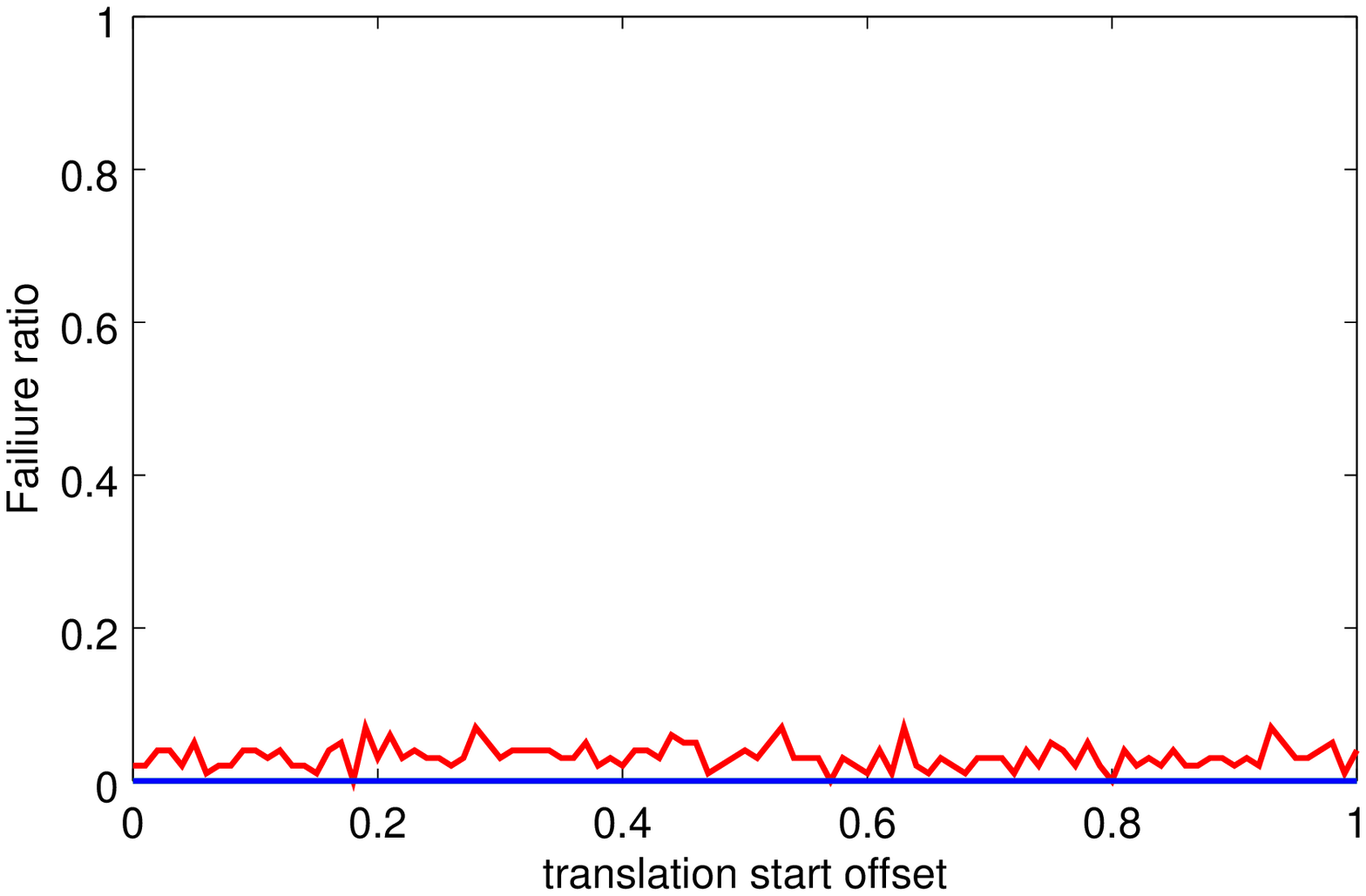}
	};
	\node[right=of tm, node distance=0cm, xshift=-1.7cm,font=\color{red}] (tr)	{
		\includegraphics[trim={1 40 3 55},clip, width=0.35\textwidth]{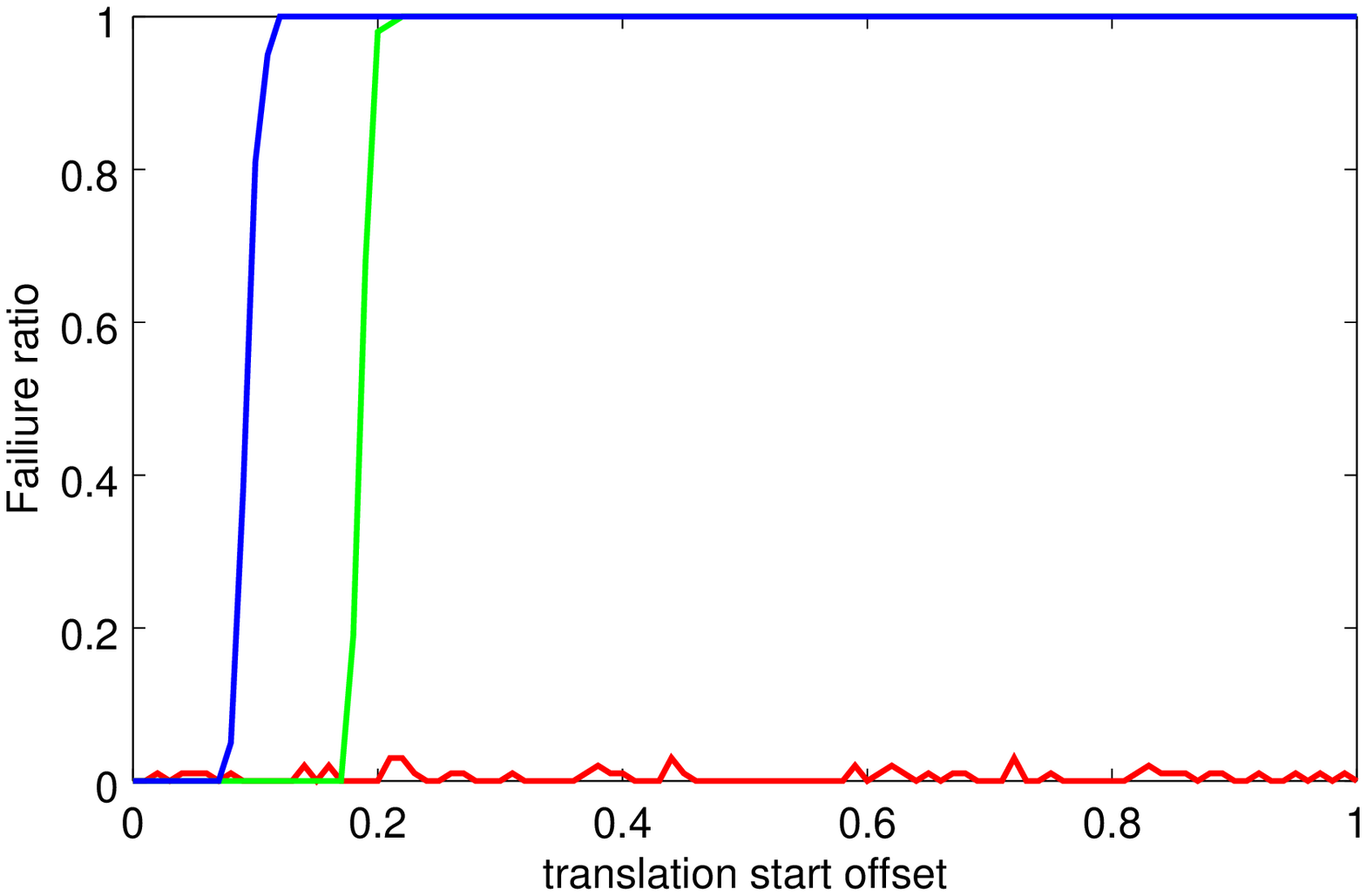}
	};

	\node[below=of tl, node distance=0cm, yshift=1.25cm,font=\color{red}] (bl) {
  		\includegraphics[trim={1 40 3 55},clip, width=0.35\textwidth]{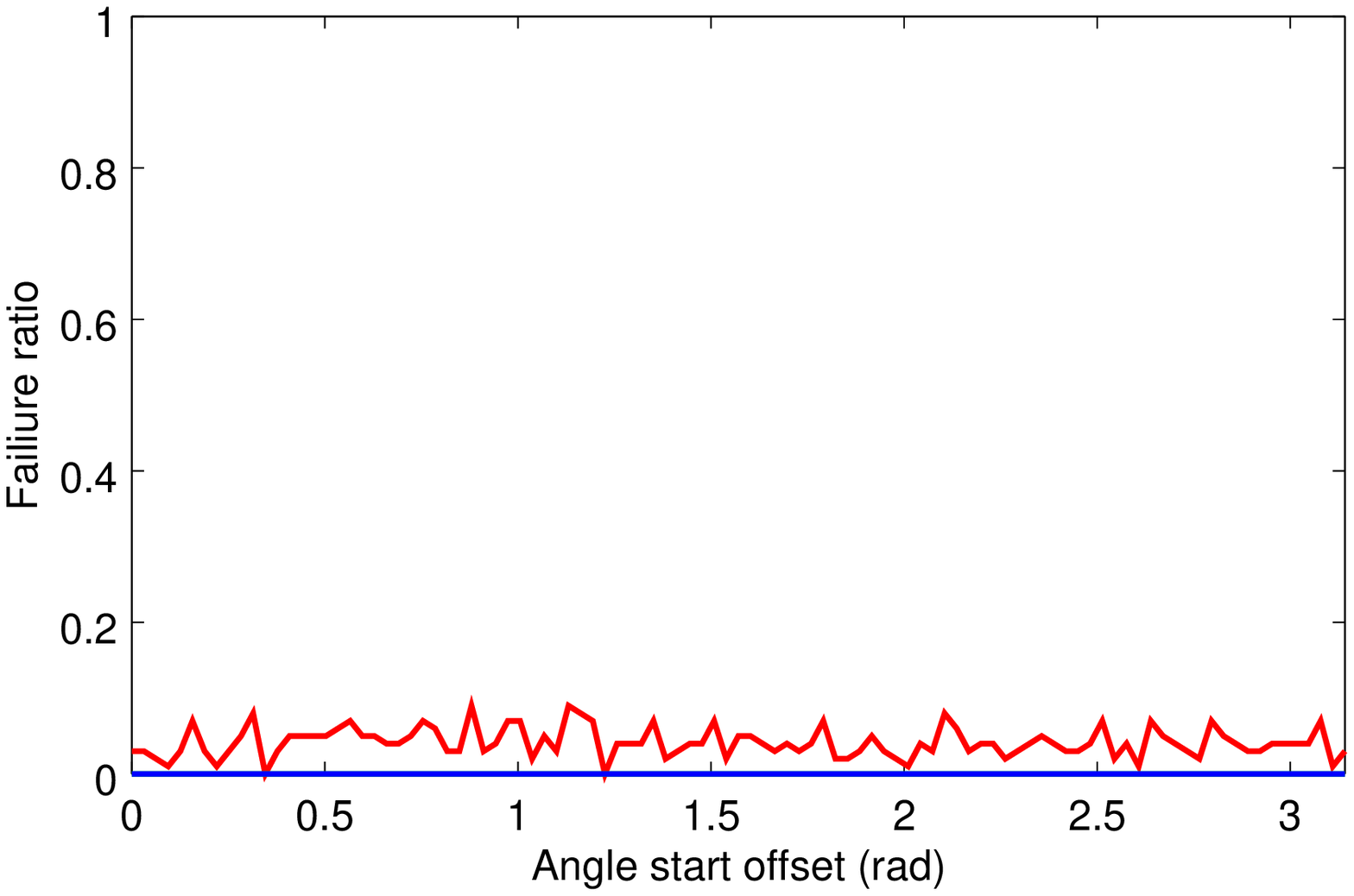}
	};
	\node[right=of bl, node distance=0cm, xshift=-1.7cm,font=\color{red}] (bm) 	{
		\includegraphics[trim={1 40 3 55},clip, width=0.35\textwidth]{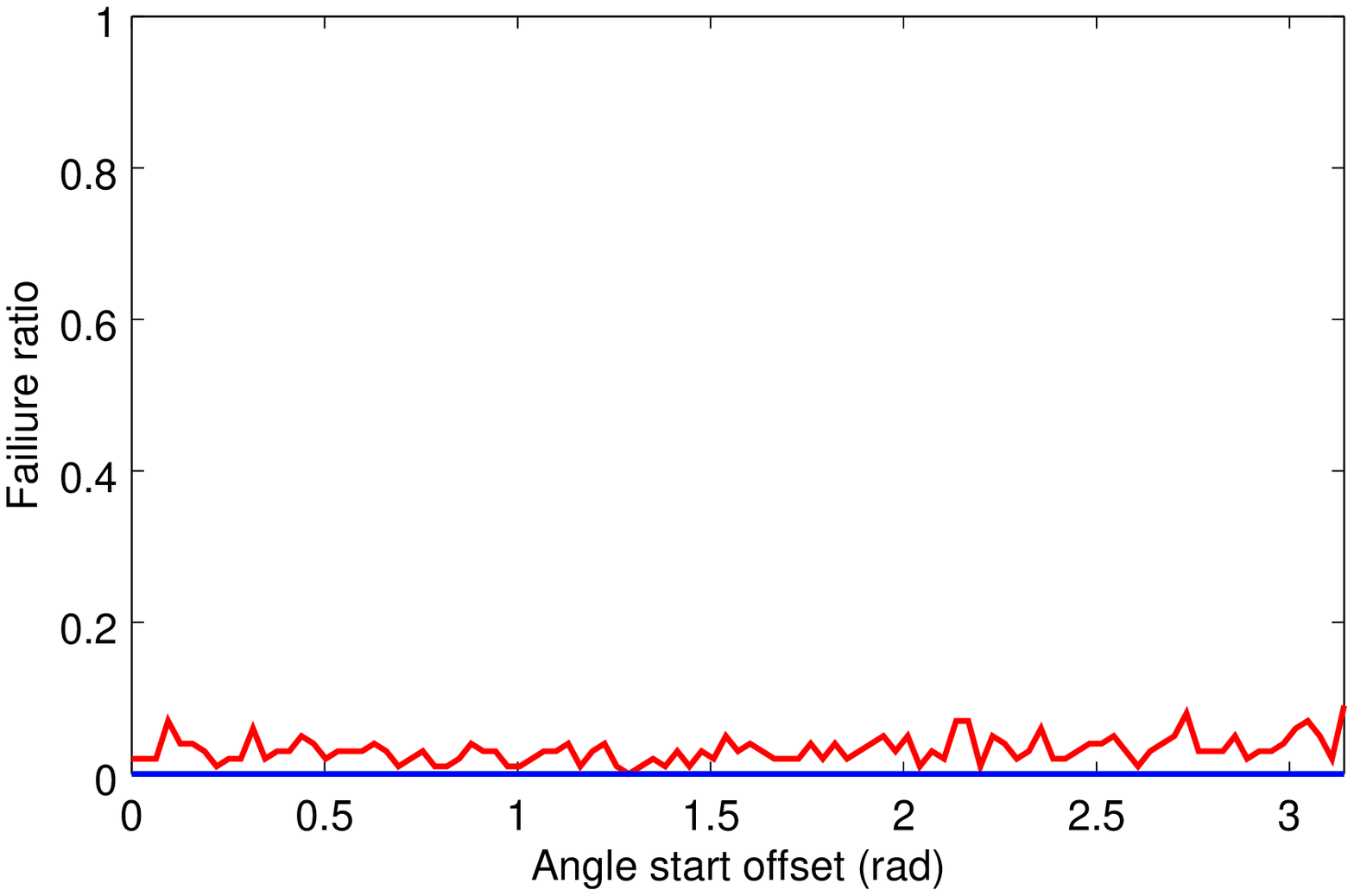}
	};
	\node[right=of bm, node distance=0cm, xshift=-1.7cm,font=\color{red}] (br)	{
		\includegraphics[trim={1 40 3 55},clip, width=0.35\textwidth]{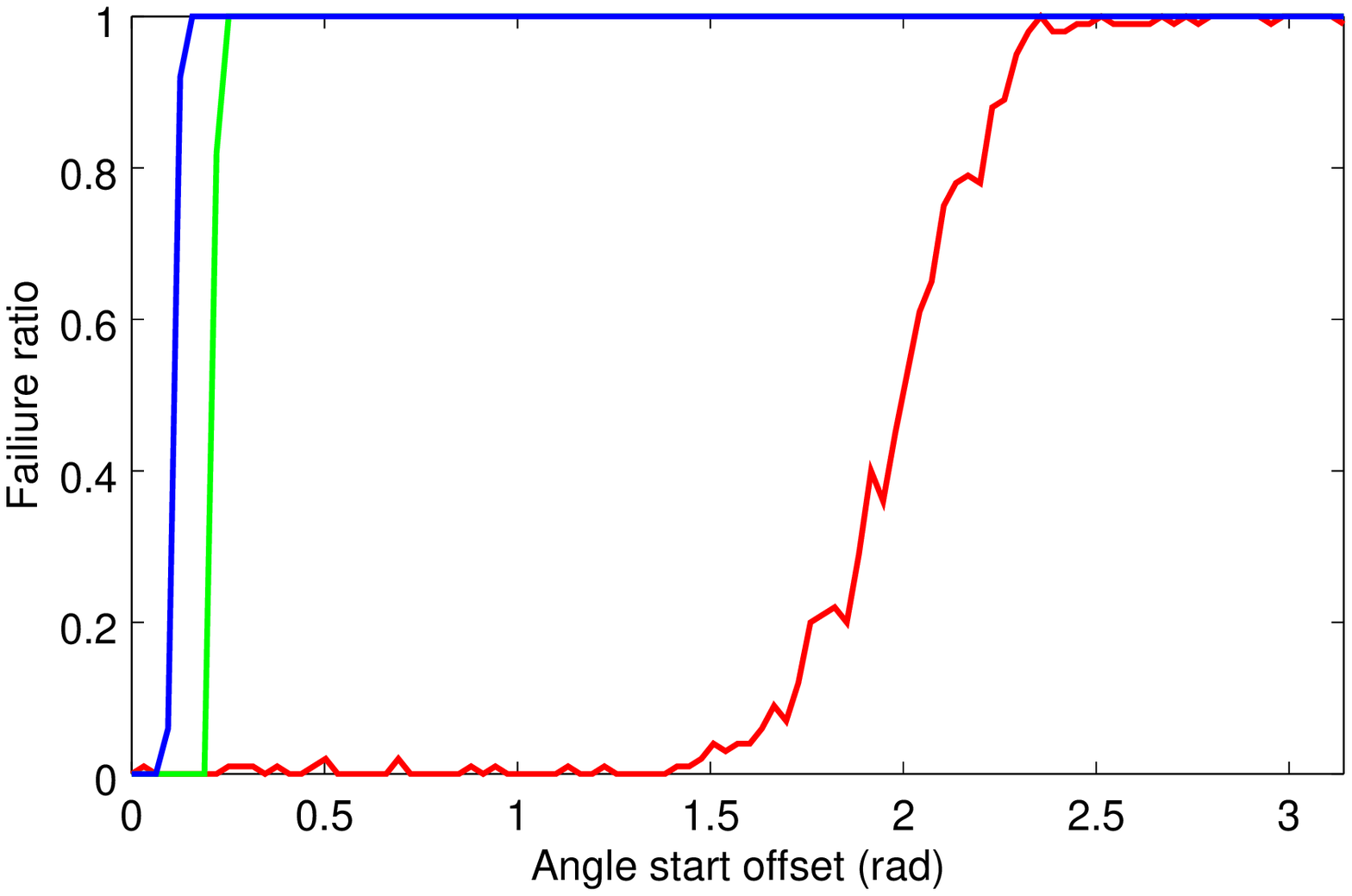}
	};	
%
  \node[above=of tl, node distance=0cm, yshift=-1.25cm,font=\color{red}]			{1000 Inliers 100 Outliers};
  \node[above=of tm, node distance=0cm, yshift=-1.25cm,font=\color{red}] (test)	{1000 Inliers 1000 Outliers};
  \node[above=of tr, node distance=0cm, yshift=-1.25cm,font=\color{red}] 		{1000 Inliers 10000 Outliers};
  \node[left=of tl, node distance=0cm, rotate=90, anchor=center,yshift=-1.0cm,font=\color{red}] {Translation};
  \node[left=of bl, node distance=0cm, rotate=90, anchor=center,yshift=-1.0cm,font=\color{red}] {Rotation};
  
  \node[above=of test, node distance=0cm, yshift=-1cm,font=\color{black}] (test2) { 
\begin{tikzpicture}
	\draw ( 0.0, 0.5) -- (0,-0.5) -- (17,-0.5) -- (17, 0.5) -- ( 0.0, 0.5);
	\draw[-][draw=black,		very thick] ( 0.0, 0)	-- ( 0.0,0) ;
	\draw[-][draw=red,		very thick] ( 1.0, 0)	-- ( 3.0,0) node[right] {  SIE L$_2$-norm};
	\draw[-][draw=green,		very thick] ( 6.0, 0)	-- ( 8.0,0) node[right] {  SIE L$_1$-norm};
	\draw[-][draw=blue,		very thick] ( 11.0, 0)	-- ( 13.0,0) node[right] {  SIE estimated norm};
\end{tikzpicture}
	};
\end{tikzpicture}
    \caption{Failure ratios for compared solutions with different number of outliers and different numbers of initial transformation estimates. The measurements are sampled with Laplacian noise. The top row contain experiments of initial transformation translated along the x-axis by 0 to 1 units. The bottom row contain experiments of initial transformation rotated around the x-axis by 0 to $\pi$ radians. In some experiments, some solutions never had any complete failures. This means that the curves overlap of along the x-axis of the figure. Therefore, if a solution is not visible in the figure, no complete failures were recorded. }
	\label{testFailLaplacian}
\end{figure*}

If $p$ is unknown, SIE can be used to find an estimate of $p$. 
Naturally this will result in reduced precision, as compared to knowing the exact value of $p$ apriori. 
In figures \ref{testErrorLaplacian} and \ref{testFailLaplacian} we perform experiments using laplacian noise, we observe that using $p = 1$ results in the best estimation, with SIE estimating $p$ outperforming the estimation when $p = 2$ when there are few to moderate amounts of outliers relative to inliers.

Both the Laplacian distribution and the estimated $p$ norm allow for fat-tail distributions, resulting in the estimated model over-fitting to the outliers if the outliers significantly outnumber the inliers.

\begin{figure*}
\begin{tikzpicture}
	\node (tl)  																{ 
		\includegraphics[trim={1 40 3 55},clip, width=0.35\textwidth]{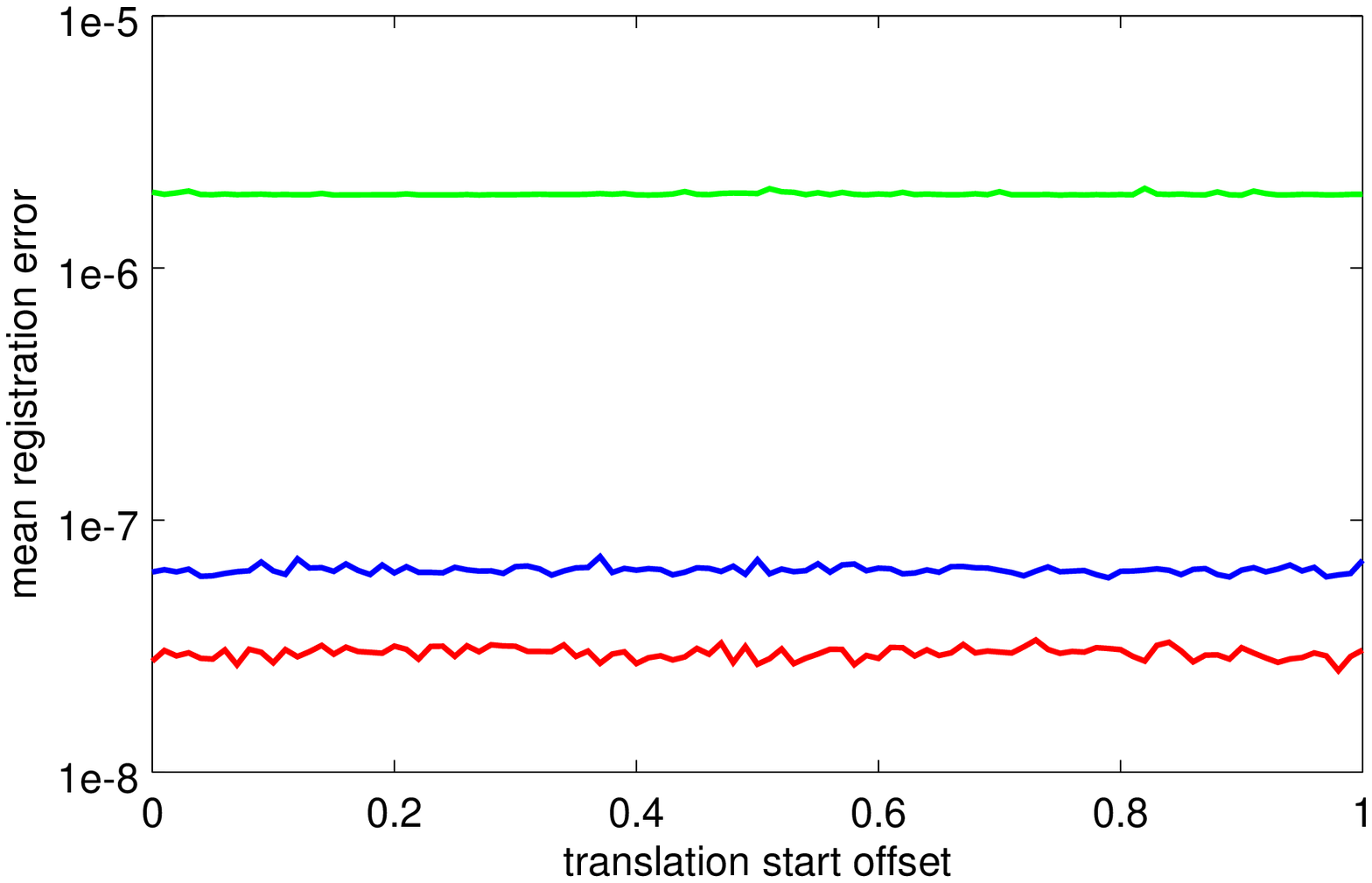}
	};
	\node[right=of tl, node distance=0cm, xshift=-1.7cm,font=\color{red}] (tm) 	{
		\includegraphics[trim={1 40 3 55},clip, width=0.35\textwidth]{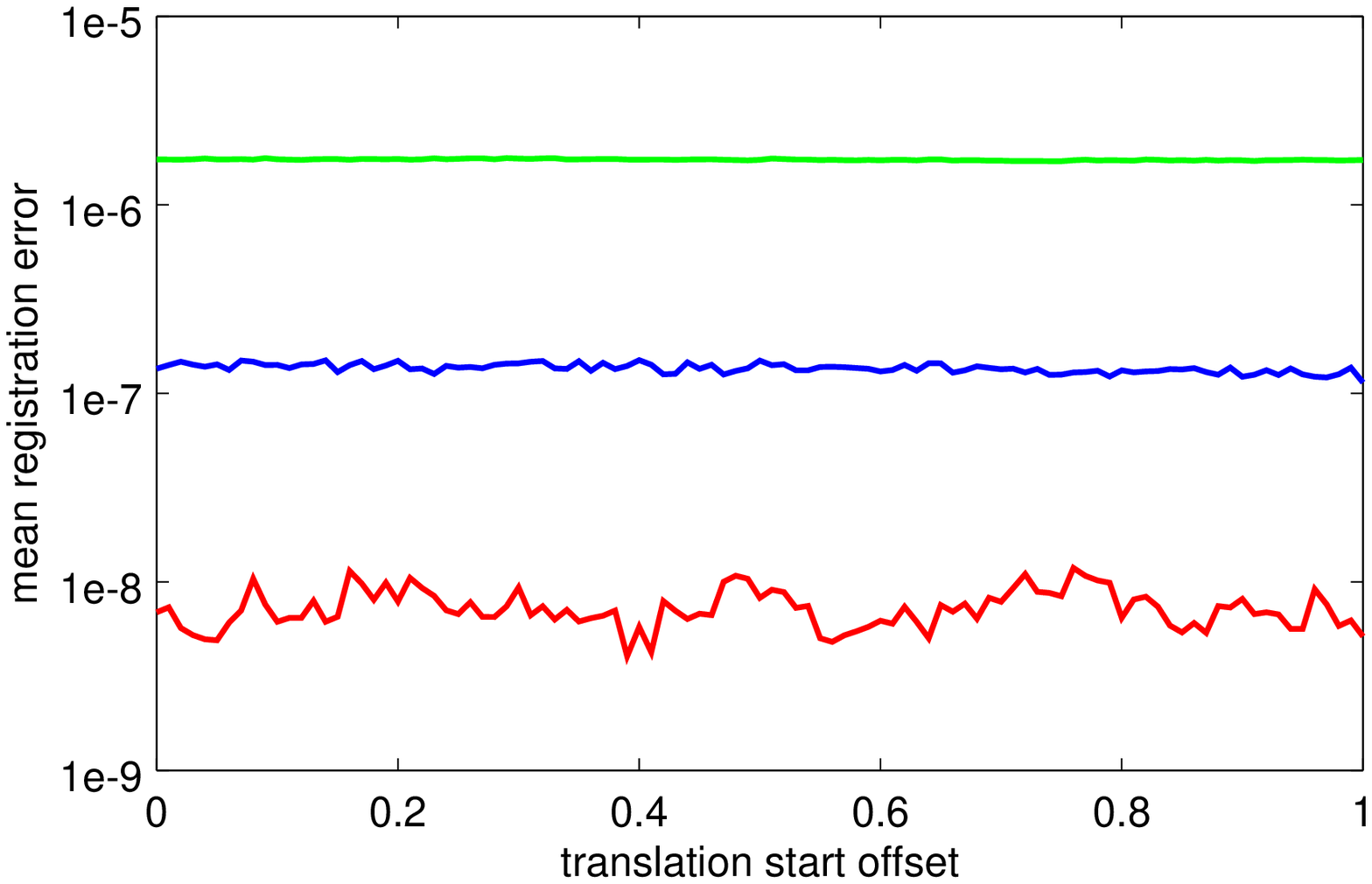}
	};
	\node[right=of tm, node distance=0cm, xshift=-1.7cm,font=\color{red}] (tr)	{
		\includegraphics[trim={1 40 3 55},clip, width=0.35\textwidth]{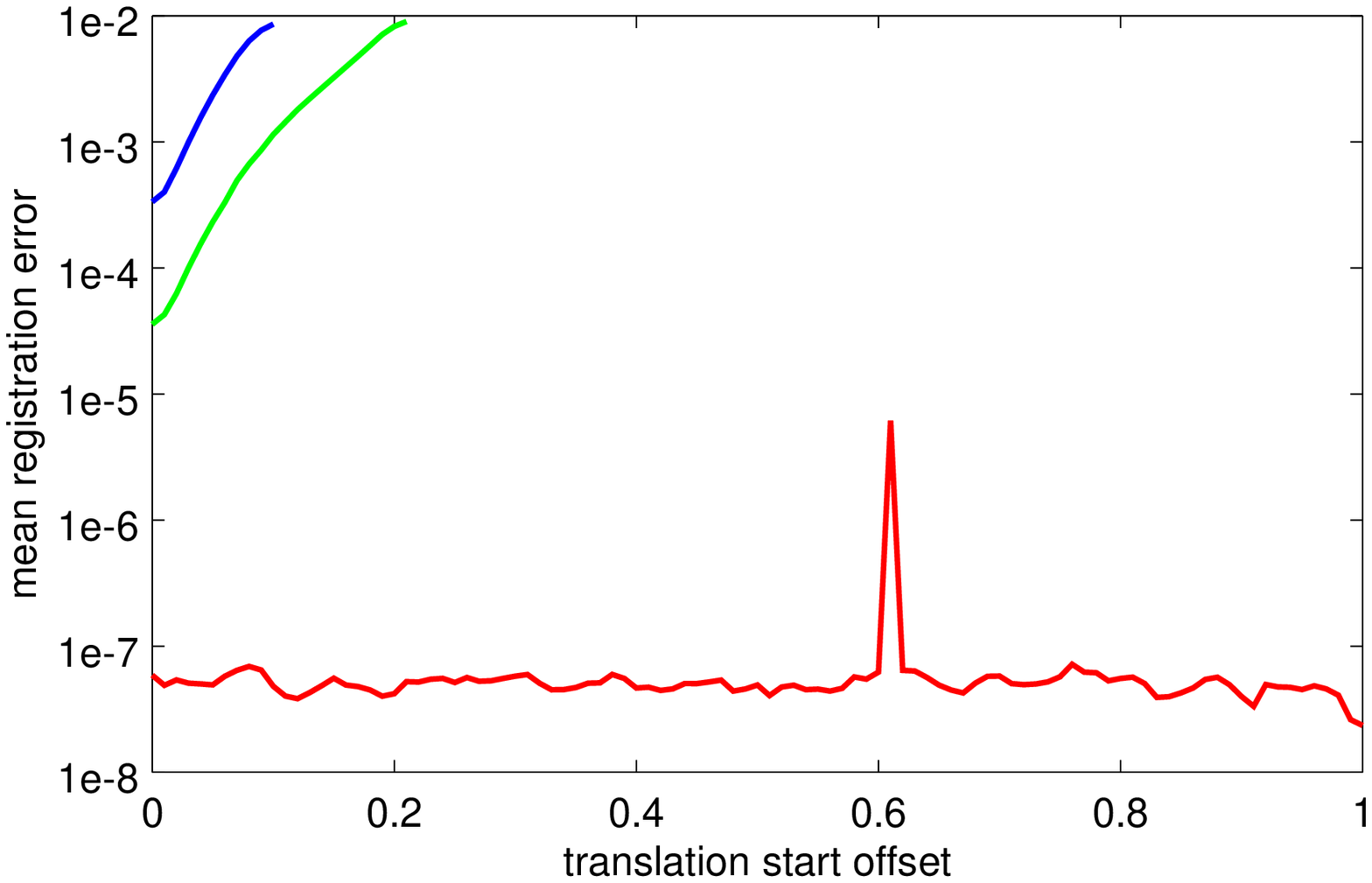}
	};

	\node[below=of tl, node distance=0cm, yshift=1.25cm,font=\color{red}] (bl) {
  		\includegraphics[trim={1 40 3 55},clip, width=0.35\textwidth]{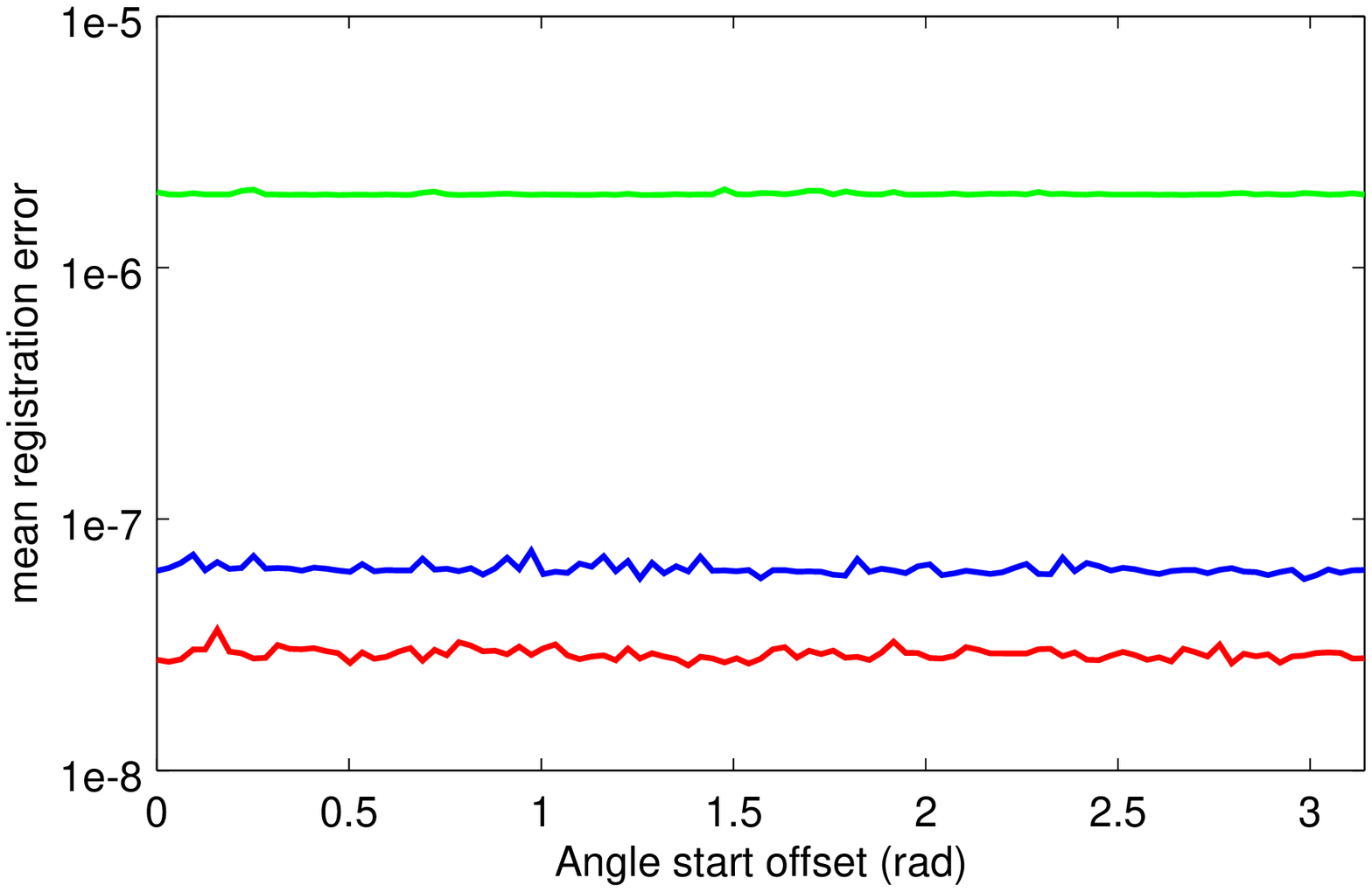}
	};
	\node[right=of bl, node distance=0cm, xshift=-1.7cm,font=\color{red}] (bm) 	{
		\includegraphics[trim={1 40 3 55},clip, width=0.35\textwidth]{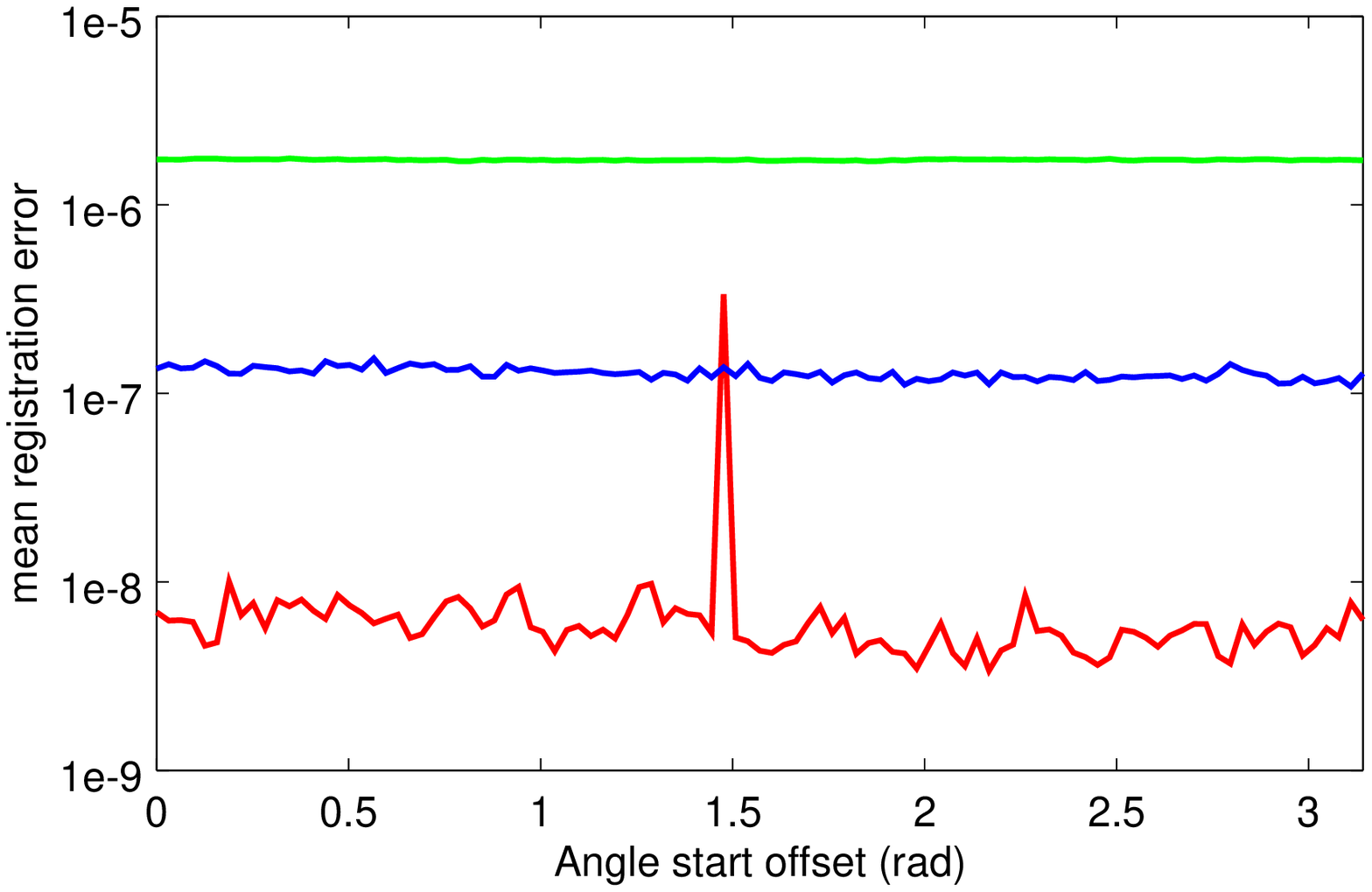}
	};
	\node[right=of bm, node distance=0cm, xshift=-1.7cm,font=\color{red}] (br)	{
		\includegraphics[trim={1 40 3 55},clip, width=0.35\textwidth]{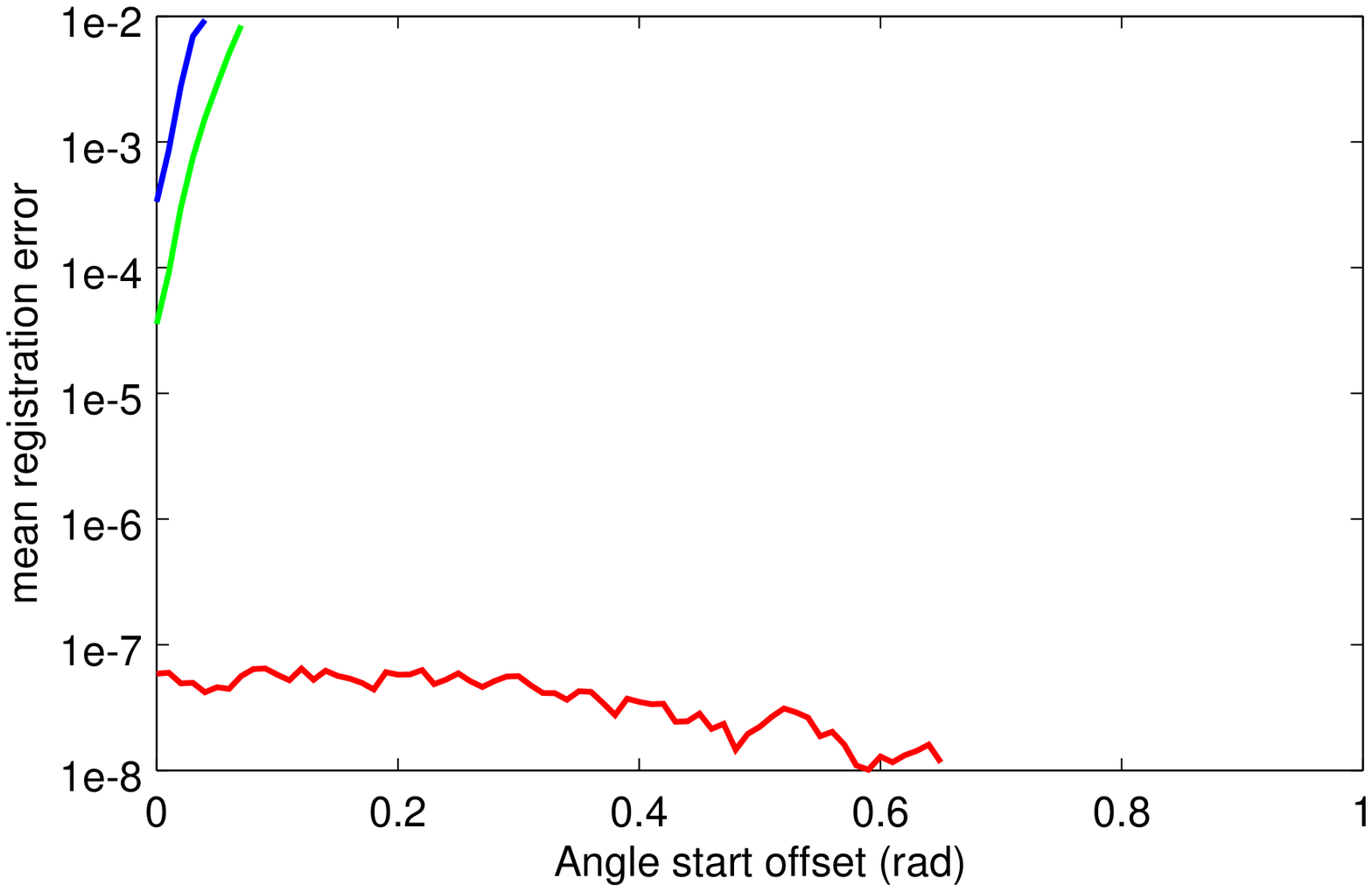}
	};	

  \node[above=of tm, node distance=0cm, yshift=0.5cm,font=\color{black}]			{\huge Gaussian noise};
  \node[above=of tl, node distance=0cm, yshift=-1.25cm,font=\color{red}]			{1000 Inliers 100 Outliers};
  \node[above=of tm, node distance=0cm, yshift=-1.25cm,font=\color{red}] (test)	{1000 Inliers 1000 Outliers};
  \node[above=of tr, node distance=0cm, yshift=-1.25cm,font=\color{red}] 		{1000 Inliers 10000 Outliers};
  \node[left=of tl, node distance=0cm, rotate=90, anchor=center,yshift=-1.0cm,font=\color{red}] {Translation};
  \node[left=of bl, node distance=0cm, rotate=90, anchor=center,yshift=-1.0cm,font=\color{red}] {Rotation};
  
  \node[above=of test, node distance=0cm, yshift=-1cm,font=\color{black}] (test2) { 
\begin{tikzpicture}
	\draw ( 0.0, 0.5) -- (0,-0.5) -- (17,-0.5) -- (17, 0.5) -- ( 0.0, 0.5);
	\draw[-][draw=black,		very thick] ( 0.0, 0)	-- ( 0.0,0) ;
	\draw[-][draw=red,		very thick] ( 1.0, 0)	-- ( 3.0,0) node[right] {  SIE L$_2$-norm};
	\draw[-][draw=green,		very thick] ( 6.0, 0)	-- ( 8.0,0) node[right] {  SIE L$_1$-norm};
	\draw[-][draw=blue,		very thick] ( 11.0, 0)	-- ( 13.0,0) node[right] {  SIE estimated norm};
\end{tikzpicture}
	};  
  
\end{tikzpicture}
    \caption{ Mean errors of non-failure cases for compared solutions with different number of outliers and different numbers of initial transformation estimates. The measurements are sampled with Gaussian noise. If the failure rate is greater than 0.5, no mean error is displayed.  The accuracy of estimation vary by orders of magnitudes between the compared solutions, the mean error is therefore drawn on a logarithmic axis. The top row contain experiments of initial transformation translated along the x-axis by 0 to 1 units. The bottom row contain experiments of initial transformation rotated around the x-axis by 0 to $\pi$ radians. }
	\label{testErrorGaussian}
\vspace{0.5 cm}
\begin{tikzpicture}
	\node (tl)  																{ 
		\includegraphics[trim={1 40 3 55},clip, width=0.35\textwidth]{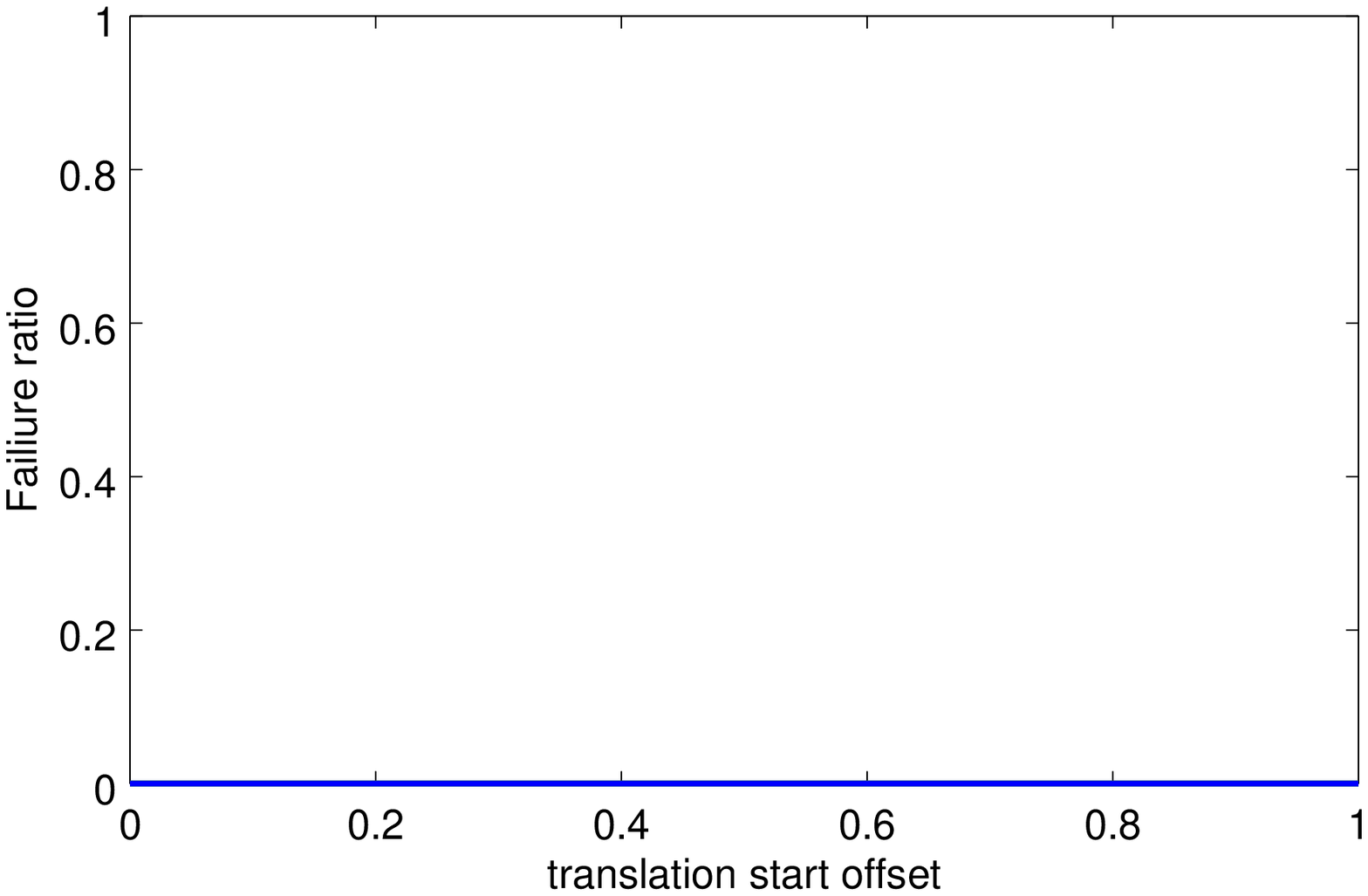}
	};
	\node[right=of tl, node distance=0cm, xshift=-1.7cm,font=\color{red}] (tm) 	{
		\includegraphics[trim={1 40 3 55},clip, width=0.35\textwidth]{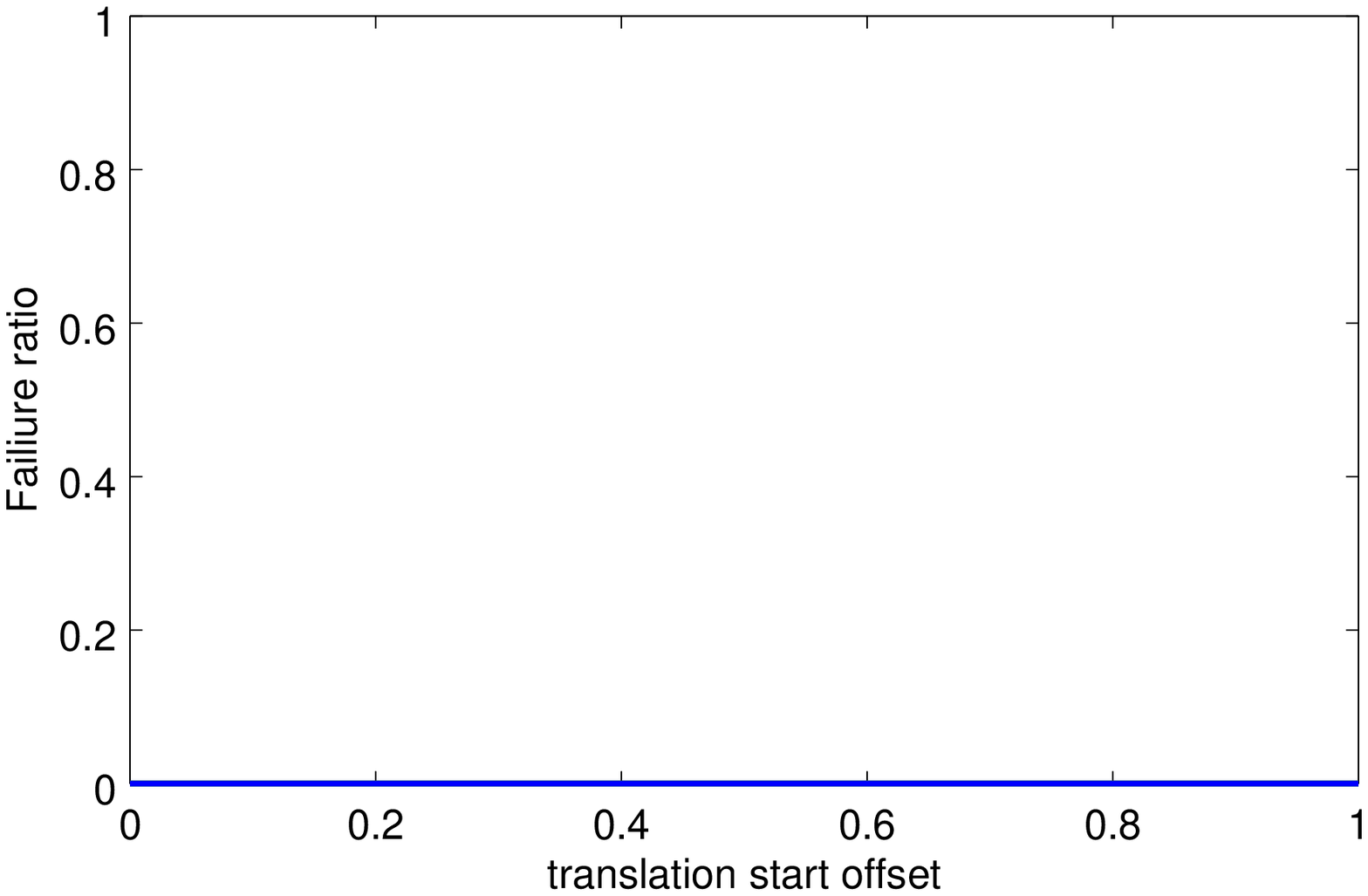}
	};
	\node[right=of tm, node distance=0cm, xshift=-1.7cm,font=\color{red}] (tr)	{
		\includegraphics[trim={1 40 3 55},clip, width=0.35\textwidth]{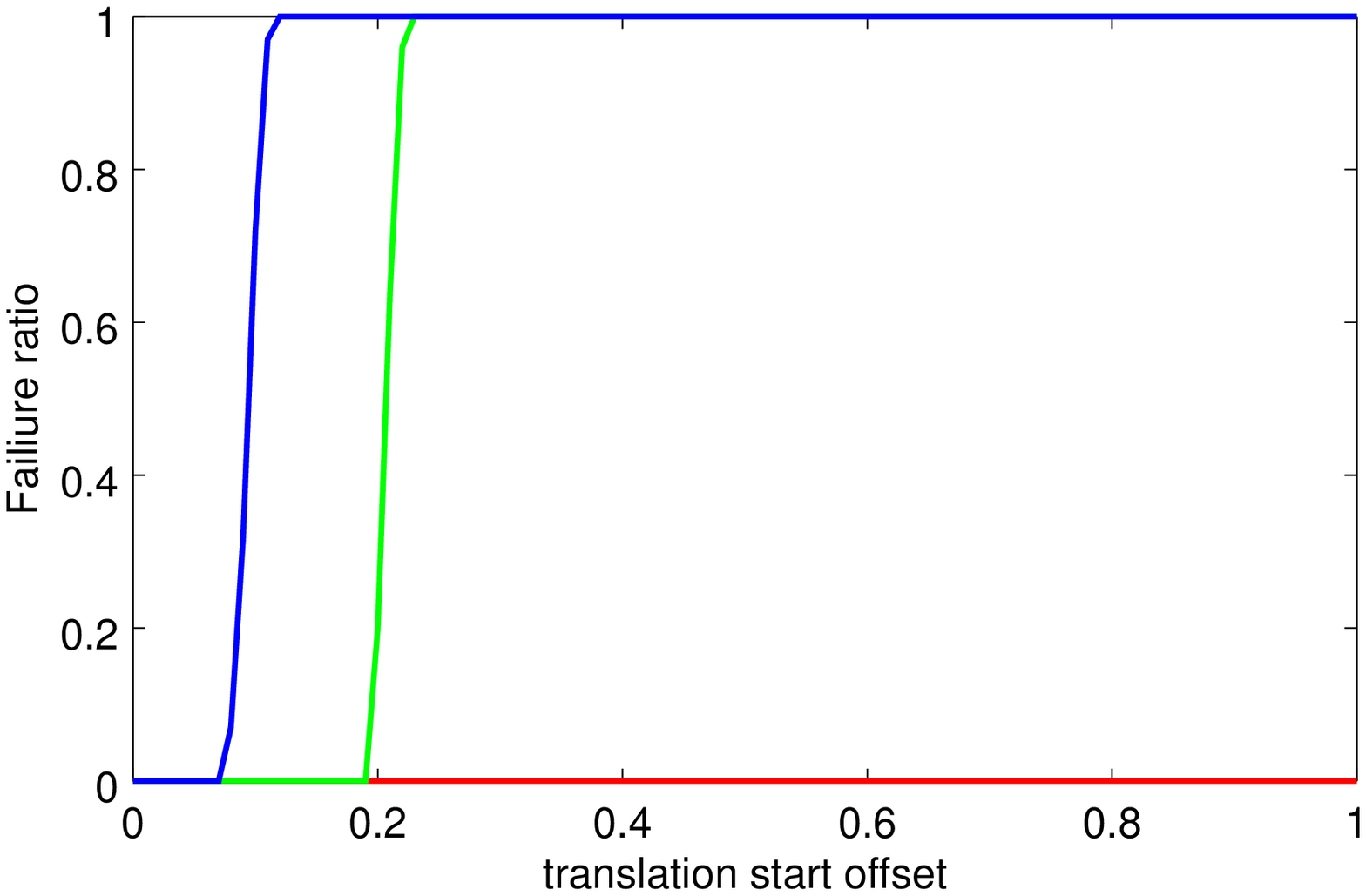}
	};

	\node[below=of tl, node distance=0cm, yshift=1.25cm,font=\color{red}] (bl) {
  		\includegraphics[trim={1 40 3 55},clip, width=0.35\textwidth]{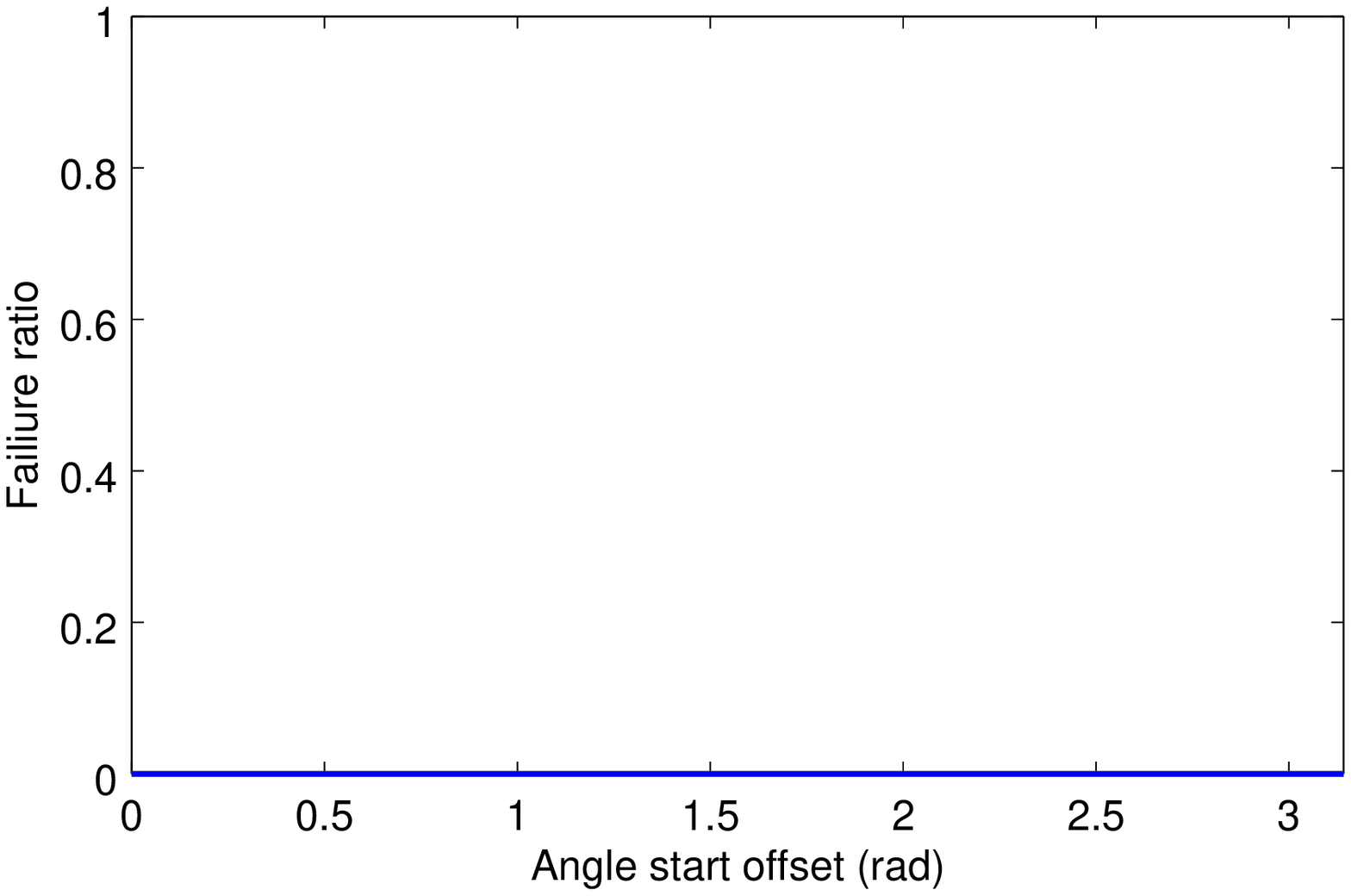}
	};
	\node[right=of bl, node distance=0cm, xshift=-1.7cm,font=\color{red}] (bm) 	{
		\includegraphics[trim={1 40 3 55},clip, width=0.35\textwidth]{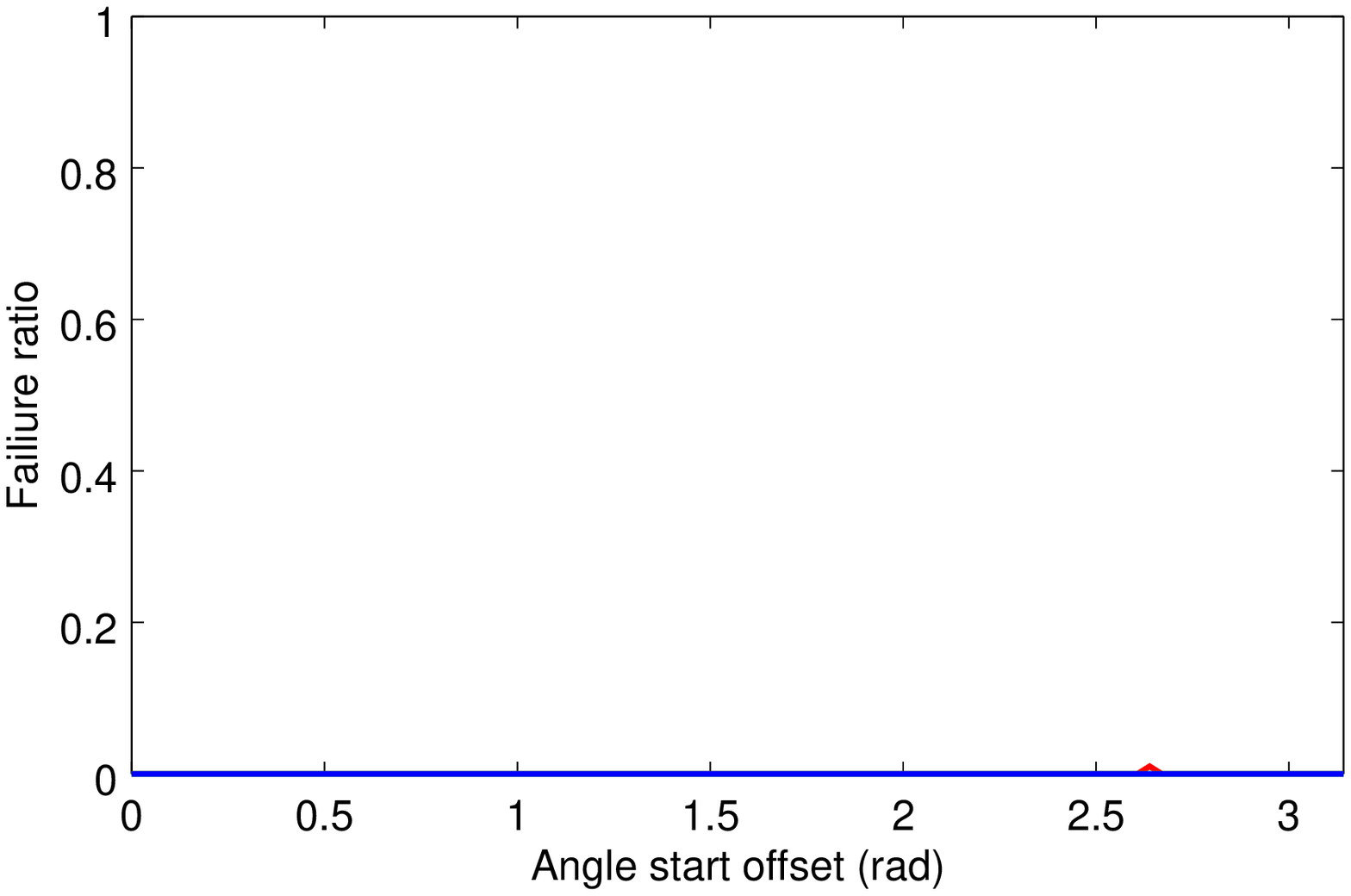}
	};
	\node[right=of bm, node distance=0cm, xshift=-1.7cm,font=\color{red}] (br)	{
		\includegraphics[trim={1 40 3 55},clip, width=0.35\textwidth]{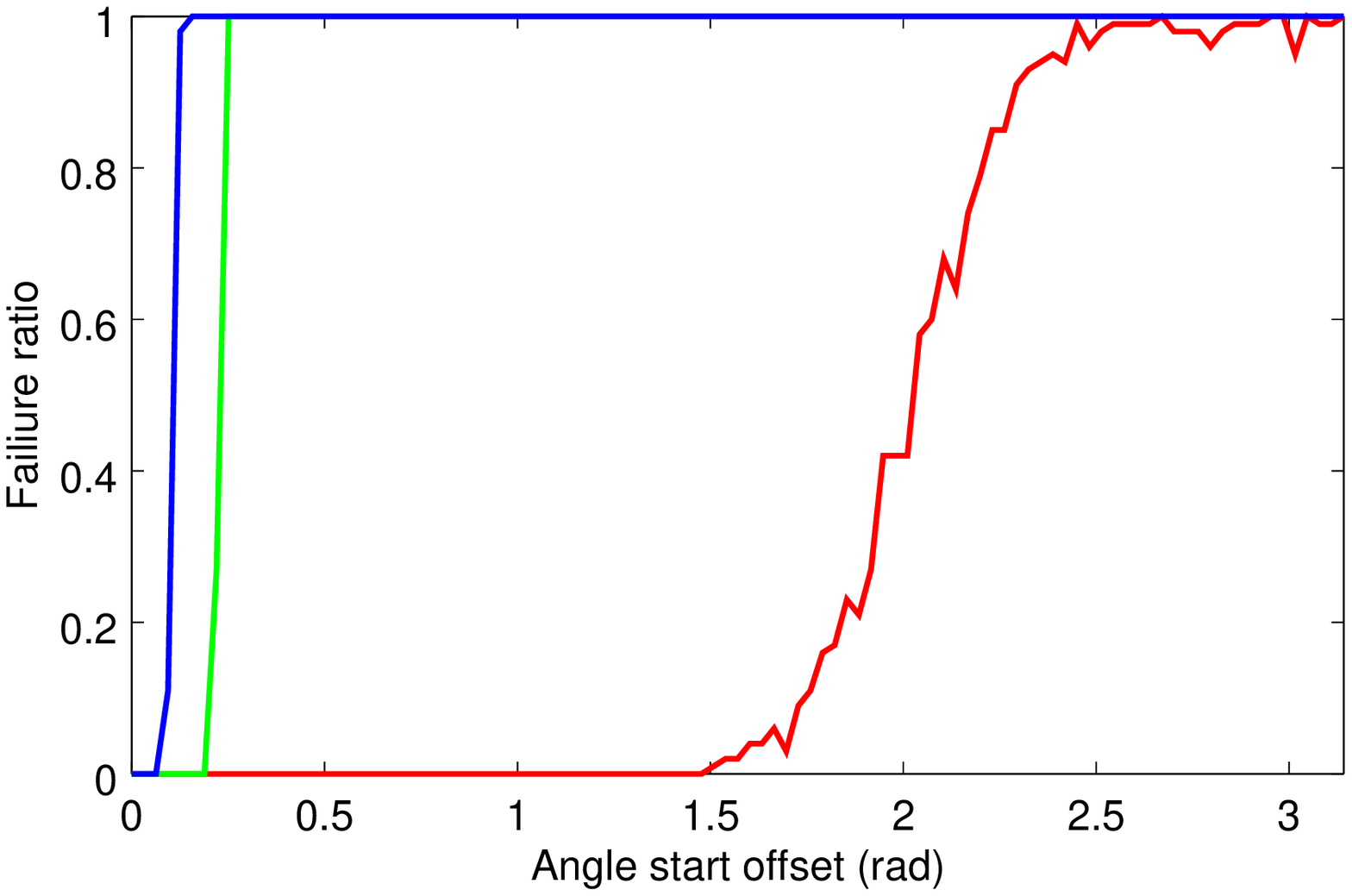}
	};	
%
  \node[above=of tl, node distance=0cm, yshift=-1.25cm,font=\color{red}]			{1000 Inliers 100 Outliers};
  \node[above=of tm, node distance=0cm, yshift=-1.25cm,font=\color{red}] (test)	{1000 Inliers 1000 Outliers};
  \node[above=of tr, node distance=0cm, yshift=-1.25cm,font=\color{red}] 			{1000 Inliers 10000 Outliers};
  \node[left=of tl, node distance=0cm, rotate=90, anchor=center,yshift=-1.0cm,font=\color{red}] {Translation};
  \node[left=of bl, node distance=0cm, rotate=90, anchor=center,yshift=-1.0cm,font=\color{red}] {Rotation};
  
  \node[above=of test, node distance=0cm, yshift=-1cm,font=\color{black}] (test2) { 
\begin{tikzpicture}
	\draw ( 0.0, 0.5) -- (0,-0.5) -- (17,-0.5) -- (17, 0.5) -- ( 0.0, 0.5);
	\draw[-][draw=black,		very thick] ( 0.0, 0)	-- ( 0.0,0) ;
	\draw[-][draw=red,		very thick] ( 1.0, 0)	-- ( 3.0,0) node[right] {  SIE L$_2$-norm};
	\draw[-][draw=green,		very thick] ( 6.0, 0)	-- ( 8.0,0) node[right] {  SIE L$_1$-norm};
	\draw[-][draw=blue,		very thick] ( 11.0, 0)	-- ( 13.0,0) node[right] {  SIE estimated norm};
\end{tikzpicture}
	};
\end{tikzpicture}
    \caption{Failure ratios for compared solutions with different number of outliers and different numbers of initial transformation estimates. The measurements are sampled with Gaussian noise. The top row contain experiments of initial transformation translated along the x-axis by 0 to 1 units. The bottom row contain experiments of initial transformation rotated around the x-axis by 0 to $\pi$ radians. In some experiments, some solutions never had any complete failures. This means that the curves overlap of along the x-axis of the figure. Therefore, if a solution is not visible in the figure, no complete failures were recorded. }
	\label{testFailGaussian}
\end{figure*}

In figures \ref{testErrorGaussian} and \ref{testFailGaussian} we observe that $p = 2$ is the best solution when the measurement noise is Gaussian. The second best solution is using SIE to estimate $p$. Assuming that $p = 1$ providing the worst estimates. We draw the conclusion that estimating $p$ is a good alternative when the exact shape of the measurement noise distribution is not precisely known.

\clearpage
\newpage
\bibliography{2017_efj_IROS}
\bibliographystyle{ieeetr}

\end{document}